\documentclass[10pt,journal,compsoc]{IEEEtran}
\usepackage{amsmath}
\usepackage{amssymb}
\usepackage{amsthm}
\usepackage{breqn}
\usepackage{subfig}
\usepackage{graphicx}
\usepackage{algorithm}
\usepackage{algorithmicx}
\usepackage{algpseudocode}
\title{Deep Transfer Learning for Brain Magnetic Resonance Image Multi-class Classification}
\author{Yusuf ~Brima,~\IEEEmembership{Member,~IEEE,}
	Mossadek~Hossain ~Kamal~Tushar,~\IEEEmembership{Member,~IEEE,}
	Upama~Kabir,~\IEEEmembership{Member,~IEEE }
	and ~Tariqul~Islam
	\IEEEcompsocitemizethanks{\IEEEcompsocthanksitem Yusuf Brima is a research student at the University of Dhaka.\protect\\
		\IEEEcompsocthanksitem Dr. Mossadek Hossain Kamal Tushar is a professor at the University of Dhaka.\protect \\
		\IEEEcompsocthanksitem Dr. Upama Kabir is a professor at the University of Dhaka.\protect \\
		\IEEEcompsocthanksitem Dr. Tariqul Islam is a research scientist at the National Institute of Neuroscience and Hospital, Dhaka, Bangladesh. }\\
	\thanks{}}
\begin{document}
	\maketitle
	\begin{abstract}
Magnetic Resonance Imaging (MRI) is a principal diagnostic approach used in the field of radiology to create images of the anatomical and physiological structure of patients. MRI is the prevalent medical imaging practice to find abnormalities in soft tissues. Traditionally they are analyzed by a radiologist to detect abnormalities in soft tissues, especially the brain. The process of interpreting a massive volume of patient's MRI is laborious. Hence, the use of Machine Learning methodologies can aid in detecting abnormalities in soft tissues with considerable accuracy. In this research, we have curated a novel dataset and developed a framework that uses Deep Transfer Learning to perform a multi-classification of tumors in the brain MRI images. In this paper, we adopted the Deep Residual Convolutional Neural Network (ResNet50) architecture for the experiments along with discriminative learning techniques to train the model. Using the novel dataset and two publicly available MRI brain datasets, this proposed approach attained a classification accuracy of 86.40\% on the curated dataset, 93.80\% on the Harvard Whole Brain Atlas dataset, and 97.05\% accuracy on the School of Biomedical Engineering dataset. Results of our experiments significantly demonstrate our proposed framework for transfer learning is a potential and effective method for brain tumor multi-classification task.
\end{abstract}

		\begin{IEEEkeywords}
			Convolutional Neural Network, Magnetic Resonance Imaging (MRI), Transfer Learning, K-Nearest Neighbour, Brain Tumor,  
	\end{IEEEkeywords}

\section{Introduction}\label{sec:introduction}
\IEEEPARstart{R}{ecent} advances in Artificial Intelligence (AI) and Machine Learning (ML) techniques are massively revolutionising healthcare as new capabilities of automation are being applied in electronic patient record analysis, radical personalization, medical image analysis, drug discovery, etc. \cite{oguzmachine}. The application AI and ML in different sectors of Healthcare is impacting its outcomes in new and profound ways. One of these outcomes is observed in Magnetic Resonance Imaging analysis \cite {litjens2017survey}. 

Magnetic Resonance Imaging is a leading modality used in radiology to study anatomical and the physiological processes of the patients. It is the prevalent medical imaging method to identify tomors in brain scans of patients. 
MRI is frequently used to provide soft-tissue contrast because of its non-invasive approach towards medical imaging \cite{liang2000principles}.
MRI images are traditionally analyzed by a radiologist to detect abnormalities of the brain. This process of interpreting huge volumes of patient MRI scans is painstakingly difficult and time-consuming \cite{Talo2019}. In applications where distinguishing between abnormal and healthy tissue is delicate, precise interpretations become imperative \cite{mohan2018mri}. Machine learning has shown considerable ability to classify, detect correctly, and segment images with precise accuracy and processing speed \cite{litjens2017survey}. In this research, we propose a framework that uses Deep Transfer Learning to perform fine-grain classification of brain MRI images. This process entails specific categorization of brain tumor from MRI scans of patients. Our proposed framework uses a Convolutional Neural Network (CNN) based on ResNet50 architecture \cite{ResNet}.

Brain MRI image classification is an area of extensive research at the intersection of Computer Vision (CV), Machine Learning, and Biomedical Imaging. Researchers have proposed novel methods that are geared toward tackling this research problem \cite{zacharaki2009classification,mohan2018mri, Summers288}. These methods range from learning the statistical characteristics of images to dimension reduction techniques to present-day Machine Learning approaches. However, these approaches thus far present inherent limitations of dataset size requirements and computational cost.Current Machine Learning approaches
require massive training datasets and extended training
time to generalize unseen samples better. Deep learning methodologies have been proposed in \cite{moritz2000whole,mallat1989theory,chaplot2006classification,el2014computer}; in their studies, they found that a better performance is achieved, though the method required substantial data sets that require massive computational requirements and training procedures. We propose a Deep Transfer Learning framework in this paper to improve further the classification performance of brain tumor using MRI images.

Our proposed approach attained a classification accuracy of 86.40\% on the National Institute of Neuroscience \& Hospitals, Bangladesh dataset, 93.80\% on the Harvard Whole Brain Atlas dataset and 97.05\% accuracy on the Southern Medical University School of Biomedical Engineering dataset.
Our experimental results indicate the proposed framework for transfer learning is robust, feasible and efficient for brain tumor classification using MRI images.

The remaining portions of the paper are organized as follows: Section 2 presents the problem statement of MRI brain tumor classification and contribution of this research. Section 3 explains the preliminaries of  Deep Transfer Learning. Section 4 the proposed methods are discussed while section 5 experimental results, and section 6 discussion and conclusion. 
\subsection{Related Work}
The task of classifying brain MRI is an area that is actively undergoing research. There has been an extensive corpus of the literature of proposed approaches amongst researchers studying brain MRI classification. These approaches are at the intersection of Computer Vision, Image Processing and Artificial Intelligence, and Machine Learning, which have been evolving over the years. The maturation of big data, compute power, and robust Machine Learning approaches have led to predominant use cases. Supervised learning algorithms \cite{uddin2019comparing} like Support Vector Machine, Artificial Neural Network, Decision Tree and Random Forest, and unsupervised learning approaches like Fuzzy C-Means, Self-Organising Map, K-Mean Clustering etc., have been pivotal to these range of use cases \cite{mohan2018mri}.

In their paper, Zacharaki et al. \cite{zacharaki2009classification} reported a multi-stage framework with a Support Vector Machine and K-Nearest Neighbour classification algorithms to first predict three glioma grades with an accuracy of 85\% and to further the predict the glioma case into either high or low grade with an accuracy of 88\%. 
Many pieces of literature exist on frameworks for solving brain tumor classification using MRI images \cite{moritz2000whole,mallat1989theory,chaplot2006classification,el2014computer}. Paul et al., in their 2017 work, presented two deep learning methods to predict brain tumor using axial MRI data. They proposed a fully connected neural network as their first method and a deep convolutional neural network as their second approach with a 91.43\% overall accuracy \cite{paul2017deep}. 

Chaplot et al., in their 2006 paper\cite{chaplot2006classification}, proposed methods using Self-Organizing Map, Artificial Neural Network, and a Support Vector Machine for brain MRI classification. El-Dahshan et al.\cite{el2014computer}, in their work, presented a novel technique consisting of Discrete Wavelet Transform, Principal Component Analysis to reduce feature space of the dataset, and a Feed-Forward Neural Network for a binary classification task of 101 normal and abnormal brain MRI images. Prior research in this direction strongly underscores the relevance of automating brain MRI classification to aid neuroscientists improve healthcare outcomes. Contrary to conventional classification methods that extensively depend on feature engineering as a precursor to the classification task, a Convolutional Neural Networks learns feature representations hierarchically and directly using the data \cite{mohan2018mri}. This approach is crucial to enabling the classifier to learn complex, rich representations of factors of variation from phenomena.
Ertosun and Robin utilized a Convolutional Neural Network pipeline comprising two modules to detect Lower Grade Glioma (LGG) vs. Glioblastoma Multiforme (GBM) and determination of tumor grade using pathological data from The Cancer Genome Atlas \cite{ertosun2015automated}. Their experimental results achieved 96\% accuracy of GBM v LGG  classification, and 71\% for the identification of the tumor grades. Paul et al., proposed two approaches using Fully Connected Neural Networks and Convolutional Neural Network for the classification of three tumor variants: meningioma, glioma, and pituitary \cite{paul2017deep}. Their results show that using axial brain images increases the classification accuracy of the neural network. 
Afshar et al., in \cite{afshar2019capsule} have proposed a Capsule Network that combines brain MRI images with coarse tumor boundaries to learn spatial relations to enable better classification performance. Their study obtained an accuracy of 90.89\%. Anaraki et al., in their 2019 paper \cite{anaraki2019magnetic}, presented an approach that combines a Convolutional Neural Network with a Genetic Algorithm for Optimization used to classify three Glioma sub-types achieving 90.0\% accuracy and in another experiment Glioma, Meningioma, and Pituitary tumor variants obtaining a 94.2\% accuracy. 
Zang et al. \cite{zhang2013mr} have presented a hybrid approach that utilized Digital Wavelet Transform as feature extractor and Principal Component Analysis for dimensionality reduction, and a Kernel Support Vector Machine with Particle Swarm Optimization algorithm for the estimation of C and $\sigma$ parameters. Saritha et al. \cite{saritha2013classification} demonstrated Wavelet Entropy-based Spider Web Plots as features coupled with a Probabilistic Neural Network to classify brain tumor images achieving 100\% accuracy. Wang et al., in \cite{wang2015feed} used a pipeline of Stationary Wavelet Transform (SWT) for feature extraction, Principal Component Analysis for dimensionality reduction combined with Particle Swarm Optimization, and Artificial Bee Colony for MRI brain images. Zhang et al. \cite{zhang2015preclinical} in their 2015 paper, proposed a novel Computer-Aided Diagnostic system that comprises Discrete Wavelet Packet Transform (DWPT), which serves as a wavelet packet coefficient extractor, Shannon and Tsallis entropies which obtained entropy features from DWPT. Finally, they used a generalized Eigenvalue Proximate Support Vector Machine with a Radial Basis Function to classify MRI brain images into normal and abnormal classes.

Nayak et al., in their 2016 work, proposed a novel classification framework the utilized 2-D discrete Wavelet Transform to extract feature vectors, Probabilistic Principal Component Analysis for feature space reduction, and AdaBoost with Random Forests classifiers. Nayak et al. \cite{ranjan2017stationary}  further proposed a Computer-Aided Diagnosis framework the employed contrast limited adaptive histogram equalization as a mechanism to enhance tumor regions in brain MRI images, 2-D stationary wavelet transform for feature extraction, AdaBoost and Support Vector Machine algorithms for normal and abnormal brain MRI classification achieving 99.45\% accuracy \cite{nayak2016brain}. Gudigar et al.,  in a comparative study assessed three multi-resolution analysis methods - Discrete Wavelet Transform, Curvelet Transform and Shearlet Transform, and Particle Swarm Optimization for textual feature extraction from transformed images using Support Vector Machine for classification \cite{gudigar2019application}. \cite{sultan2019multi} have proposed a Deep Neural Network model using Convolutional Neural Network to classify three tumor variants (glioma, meningioma and pituitary tumor) and three glioma sub-types ( Grade II, Grade III and Grade IV) achieving an overall performance accuracy of 96.13\% and 98.7\% respectively.

Deep learning (DL) is becoming a cornerstone to a great many applications across disciplines \cite{plawiak2018novel,yildirim2017texture,plawiak2014estimation,rzecki2018application}. Back-propagation algorithm has enabled DL models to construct hierarchies of concepts and representations through layered non-linear transforms with minimal feature engineering and domain knowledge \cite{bengio2015editorial}. Various DL architectures are applied increasingly to problems across disciplines, from the reconstruction of satellite imagery to drug discovery and protein structure/function prediction. 
The ability to learn complex hierarchies of factors of variability in signals is a focal reason for the surge in Deep Learning research and practice \cite{cao2018deep}. Deep Learning models are applied to a range of use cases across domains and disciplines \cite{yildirim2018arrhythmia,acharya2017deep,kiranyaz2015real,yildirim2018novel,acharya2017automated,yildirim2018efficient,yildirim2018deep}.
\section{ Problem Statement and Contribution}
\subsection{Problem Statement}
The field of neuroscience is traditionally tasked with analyzing MRI data for the detection of tumors. The procedure, however, grapples with a considerable challenge and requires a substantial level of domain expertise through extensive formal skills acquisition. Researchers have proposed various methods to address this classification task using feature selection and reduction techniques as a precursor to classification \cite{mallat1989theory,chaplot2006classification,el2014computer}. In our research, we propose a framework using  Deep Transfer Learning as a mechanism to solve the multi-class classification task of brain MRI images. Our research uses a Deep Residual CNN based on the ResNet50 architecture.

\subsection{Contributions}
Our main contribution in research is a novel end-to-end Deep Transfer Learning framework using a ResNet50 CNN architecture that performs multi-classification of brain MRI images. We aim to investigate the transferability of source domain invariant knowledge to the target domain to reduce the algorithmic training time in conjunction with improving the generalization performance of the trained model using minimal target domain dataset. 
Our second contribution is a publication of a novel brain MRI dataset of 130 patients consisting of 5285 images belonging to 37 categories in collaboration with the National Insitute of Neuroscience \& Hospitals (NINS) in Bangladesh.
In our study, we conducted experiments with our proposed NINS dataset and two public benchmark datasets: the Biomedical School of Engineering brain tumor dataset \cite{Cheng2017} that consists of 3064 T1-weighted MRI images, and the Whole Brain Atlas that contains 1133 T2-weighted Harvard Medical School benchmark dataset \cite{Summers288}.

\section{Preliminaries} 
\subsection{Transfer Learning}\label{mmethod} 
\subsubsection{Formal definition and notation} 
The framework for transfer learning comprise \textbf{domain}, \textbf{task}, and \textbf{marginal probabilities} \cite{pan2009survey,PMID:27712798} defined as follows: 
A domain, $D$, is a two-element tuple that consists of input space indicated by $\chi$, and a marginal probability, $P(X)$, $X$ is a sample data point. Thus, we can represent the domain formally as 
\begin{equation} 
	D = \{\chi,P(X)\} 
\end{equation} 
A Domain consist of two elements: $D = \{\chi, P(X)\}$ 
where: 
\begin{itemize} 
	\item Input Space: $\chi$ 
	\item Marginal distribution: 
	\begin{equation} 
		P(X), X = \{x_1,x_2,x_3, \cdots ,x_n\},\forall{ x_i}\in \chi 
	\end{equation} 
\end{itemize} 
Hence $x_i$ represents a specific vector as represented in the above depiction. A task, $T$, on the other hand, is a two-element tuple of the label space, $\gamma$, and target function, $f(.)$. From a probabilistic standpoint, the target function can be stated thus: $P(\gamma|X)$ as a conditional probability. 
Given a domain $D$, the definition of a Task $T$ is thus stated by two elements:
\begin{equation} 
	T = \{\gamma,P(Y/X)\} = \{\gamma,f(.)\} Y = \{y_1,y_2,y_3, ..., y_n\},,\forall{ y_i}\in \gamma 
\end{equation}

\begin{itemize} 
	\item A Label Space: $\gamma$ 
	\item A predictive function $f(.)$, learned from the feature vectors/label pairs $(x_i,y_i), x_i$$\in$$X,\forall{y_i}$$\in$$\gamma$ 
	
	\item For each feature vector in the domain, $f(.)$ predicts its corresponding label: $f(x_i) \equiv y_i$ 
	
\end{itemize} 
The formulation of a Transfer Learning setting where $D_S$ is the source domain, $T_S$ is the source task, $D_T$ is the target domain, and $T_T$ is the target task is thus:
\begin{equation} 
	D_S = \{\chi_S,P(X_S)\}, X_S= \{x_{S_1},x_{s_2},x_{S_3},...x_{S_n}\}\forall{x_{S_i}}\in \chi_S 
\end{equation} 

\begin{equation} 
	T_S = \{\gamma_S,P(Y_S|X_S)\}, Y_S= \{y_{S_1},y_{s_2},y_{S_3},...y_{S_n}\}\forall{y_{S_i}}\in \gamma_S 
\end{equation} 

\begin{equation} 
	D_T = \{\chi_T,P(X_T)\}, X_T= \{x_{T_1},x_{T_2},x_{T_3},...x_{T_n}\}\forall{x_{T_i}}\in \chi_T 
\end{equation} 

\begin{equation} 
	T_T = \{\gamma_T,P(Y_T|X_T)\}, Y_T= \{y_{T_1},y_{T_2},y_{T_3},...y_{T_n}\}\forall{y_{T_i}} \in \gamma_T 
\end{equation} 
Considering a classification task $T$, with $X$ as the input space and $Y$ as the set of labels. 
Given two sets of samples drawn from both the source and target domains: 
\begin{equation} 
	D_S = \{x_i,y_i\}_{i=1}^{n} \sim P(X_S) 
\end{equation} 
\begin{equation} 
	D_T = \{x_i,y_i\}_{i=n+1}^{N} \sim P(X_T) 
\end{equation} 
The goal of Transfer Learning is to build a functional mapping $f(.):X \rightarrow Y$ with a low target classification error on the target domain assuming the 
$D_S$ and $D_T$ are \textbf{i.i.d.} \\ 
$R_{D_T} = Pr(f(x)\ne y) \leftarrow $ probability of miss-classification $(x,y) \sim D_T$. Therefore, Transfer Learning as follows is defined as:
given a source domain $D_S$ with a source task $T_S $,and a target domain $D_T$ with its target task $T_T$, 
the objective of Transfer Learning is to allow the transferability of target conditional probability distribution $P(T_T|D_T)$ in $D_T$ using latent domain invariant knowledge from $D_S$ and $T_S$ given $D_S \ne D_T$ or $T_S \ne T_T$. Transfer Learning is well suited to cases where the number of target labels is disproportionately smaller than the number of source labels \cite{tan_sun}. 
\begin{figure}[h] 
\centering 
\includegraphics[width=8.5cm,height=4.5cm]{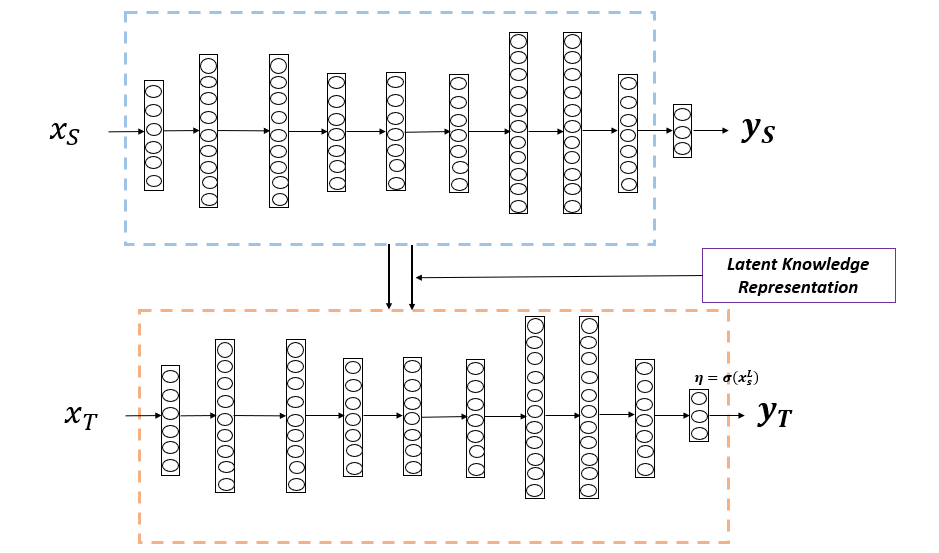} 
\caption{High-level formal representation of Transfer Learning} 
\label{fig:transfer_learning} 
\end{figure}
\section{Proposed Method}
\subsection{System Architecture}
\begin{figure}[h]
	\centering
	\includegraphics[width=8.5cm,height=7cm]{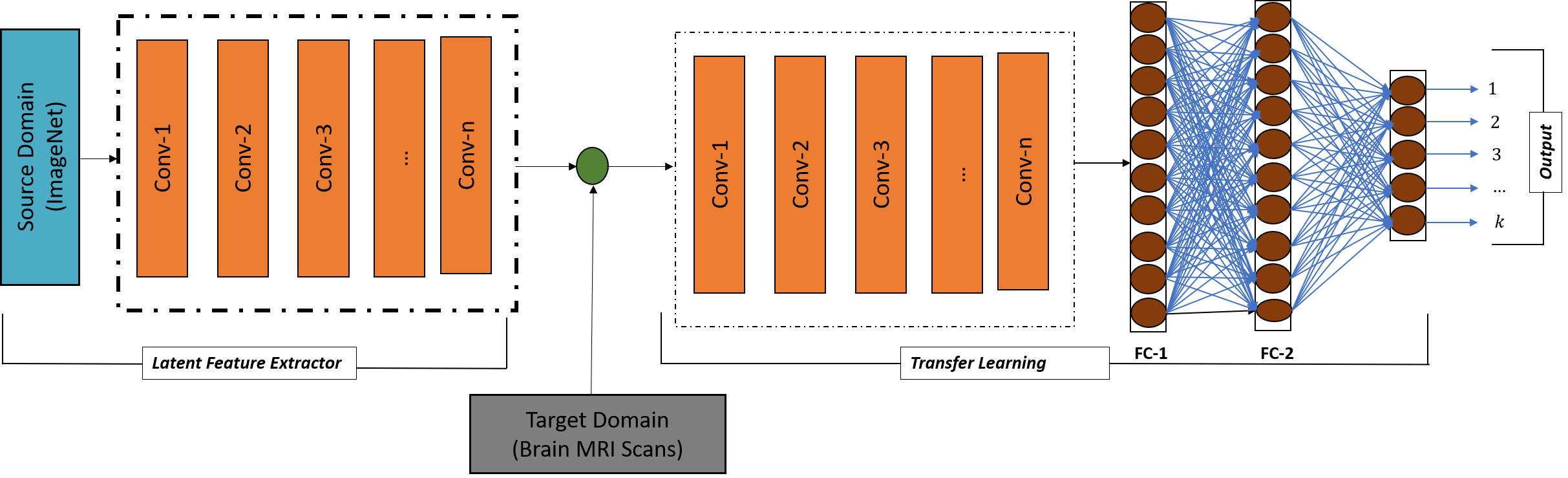}
	\caption{Proposed System Architecture}
	\label{fig:system_architecture}
\end{figure}

In this paper, we used ImageNet \cite{deng2009imagenet} as the source domain, a dataset with over one million images in one thousand categories of concepts from which the model learned and transferred invariant factors of variation to the target domain Brain MRI. Fig. \ref{fig:system_architecture} shows the architectural framework for this Representational Transfer Learning task. Transfer learning is a well-suited framework for healthcare computer vision tasks where target domain datasets for learning are significantly small, and model generalisation is a key consideration \cite{pan2009survey}.
\subsection{System Model and Assumptions}
\begin{figure}[h]
	\centering
	\includegraphics[width=8.5cm,height=4cm]{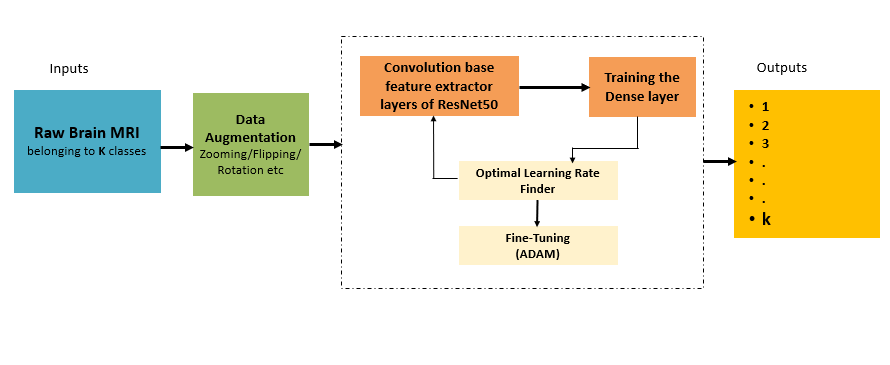}
	\caption{A flow diagram of our proposed transfer learning framework}
	\label{fig:systemdiagram}
\end{figure}

We have shown a high-level overview of our proposed Transfer Learning framework in Fig. \ref{fig:systemdiagram}. The initial stage entails the setup of the dataset which includes loading both the brain MRI image data and associated class labels followed by batch normalization and cross-validation split into train and validation sets. 
We have used various data augmentation approaches to overcome overfitting by creating virtual copies of brain MRI images. This includes methods such as zooming, flipping, rotation, mirroring etc. to up-sample the datasets. Afterwards, we used ResNet50 CNN architecture, Optimal Learning Rate Finder, Gradient Descent with Restart (SGDR), and Adaptive Moment Estimation (ADAM). We trained the model through the three stages presented in Fig. \ref{fig:system_training} accordingly.
\subsubsection{Mechanism of Deep Transfer Learning Model }
This work proposes a Transfer Learning approach to the fine-grain brain MRI classification problem using state-of-the-art pre-trained ResNet50 as the CNN architecture due to its robustness to the exploding gradient problem \cite{ResNet}. 
We have used a pre-trained network on the ImagetNet, a dataset that comprises over a million images in a thousand classes of concepts. ResNet architecture has applications in a variety of Computer Vision tasks due to its faster convergence speed relative to Inception and Visual Geometric Group, AlexNet. The architecture is very robust, simple, and useful.
\subsubsection{Proposed stages of Deep Transfer Learning}
\begin{figure}[h]
	\centering
	\includegraphics[width=8.5cm,height=6.0cm]{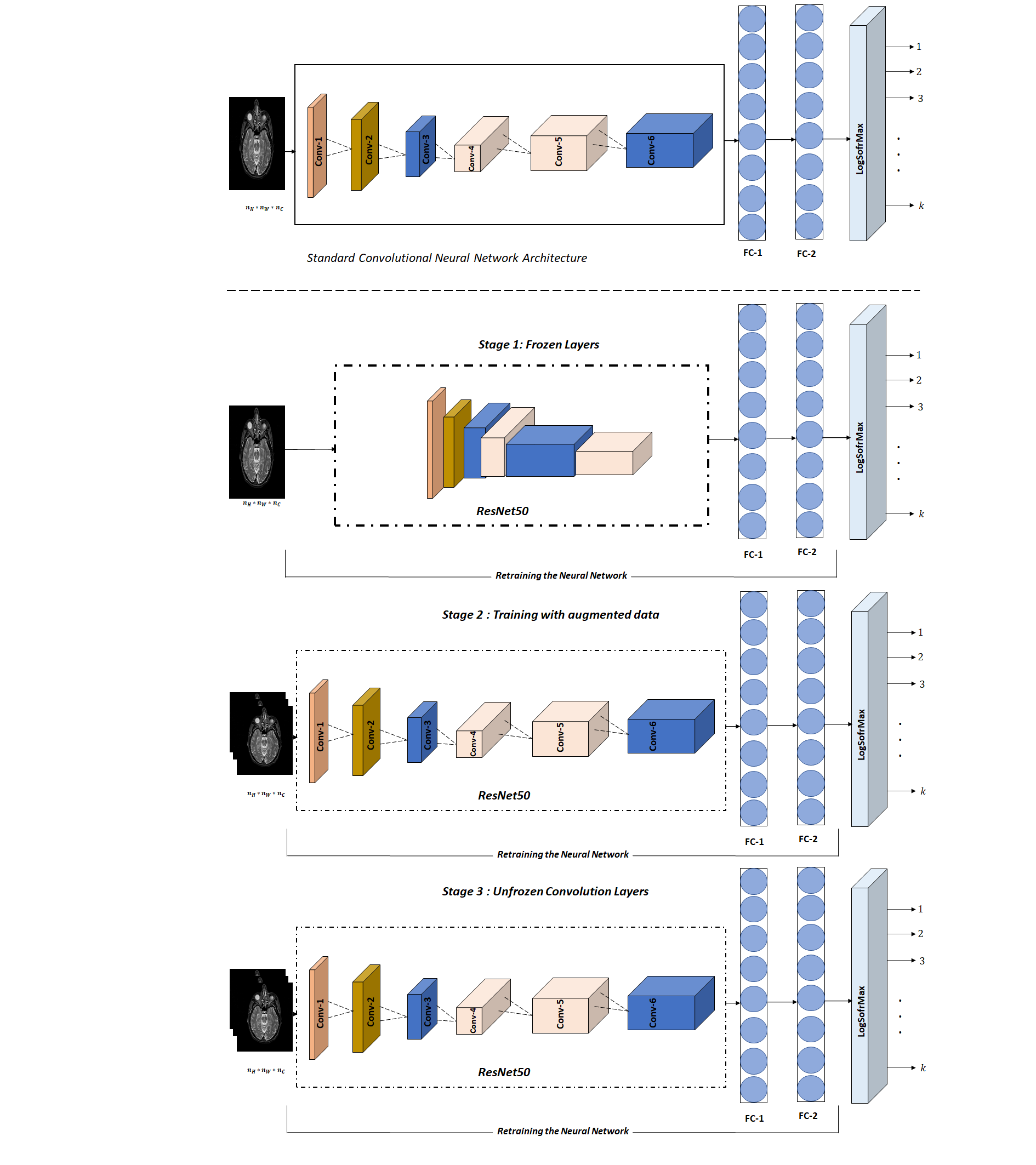}
	\caption{Deep Transfer Learning Stages 1: Retraining ResNet50; 2: Retraining the frozen layers of the network with augmented data; 3:Unfreezing and retraining the network with augmented data}
	\label{fig:system_training}
\end{figure}

The framework of Deep Transfer Learning overcomes the problem of lack of labelled data, for example, in medical imaging applications by utilizing domain-invariant tacit knowledge to solve related problems. This approach towards learning can lead to better generalization of models with minimal target domain data. Fig.\ref{fig:system_training} illustrates the three stages of our proposed transfer learning framework.
\subsection{Dataset}

In this paper, we have published a dataset consisting of 5285 T1-weighted MRI images belonging to 37 categories in collaboration with the National Insitute of Neuroscience \& Hospitals (NINS) of Bangladesh. Besides that, we have used two other public datasets, the School of Biomedical Engineering brain MRI dataset that comprises 3064 T1-weighted contrast-enhanced images of 233 study patients indicated in Table\ref{tab:biomedical_dataset} \cite{Cheng2017}. The second dataset is the  Harvard Medical School Whole Brain Atlas brain MRI  benchmark \cite{Summers288} that consists of 1133 T2-weighted brain MRI images of 38 patients shown in Table \ref{tab:harvard_medical_dataset}. The size of all images is 256x256 pixels in the axial plane.
\begin{figure}[h]
	\centering
	\includegraphics[width=8.5cm,height=10cm]{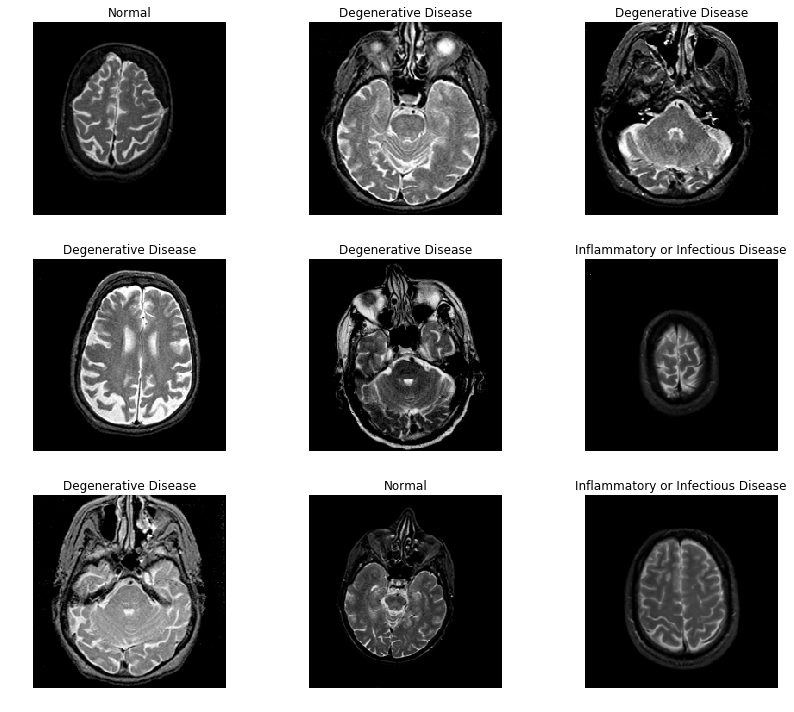}
	\caption{Sample Harvard Whole Brain MRI Dataset}
\end{figure}
The School of Biomedical Engineering, Southern Medical University brain MRI dataset comprises of three classes of brain tumors,  Meningioma, Glioma, and Pituitary tumor, as depicted in Table \ref{tab:biomedical_dataset} \cite{Cheng2017}. The Harvard Medical School dataset entails five classes, Cerebrovascular class (stroke), Neoplastic, Degenerative, and Inflammatory disease, as shown in Table \ref{tab:harvard_medical_dataset}. We have used a 5-fold cross-validation technique to perform the network re-training, which ensures that we overcome the size limit of our dataset without over-fitting.

\begin{table}[h!]
\tiny
\begin{tabular}{|l|l|l|}
\hline
Category & Total Patients & Total Slices \\ \hline
Brain Atrophy & 8 & 264 \\ \hline
Brain Infection & 2 & 38 \\ \hline
Brain Infection with abscess & 2 & 76 \\ \hline
Brain Tumor & 3 & 76 \\ \hline
Brain Tumor (Ependymoma) & 1 & 36 \\ \hline
\begin{tabular}[c]{@{}l@{}}Brain Tumor (Hemangioblastoma  \\ Pleomorphic xanthroastrocytoma  metastasis)\end{tabular} & 2 & 74 \\ \hline
Brain tumor (Astrocytoma Ganglioglioma) & 1 & 38 \\ \hline
Brain tumor (Dermoid cyst craniopharyngioma) & 1 & 38 \\ \hline
Brain tumor - Recurrenceremnant of previous lesion & 3 & 114 \\ \hline
Brain tumor operated with ventricular hemorrhage & 1 & 76 \\ \hline
Cerebral Hemorrhage & 1 & 36 \\ \hline
Cerebral venous sinus thrombosis & 1 & 76 \\ \hline
Cerebral abscess & 1 & 36 \\ \hline
demyelinating lesions & 1 & 38 \\ \hline
Encephalomalacia with gliotic change & 2 & 76 \\ \hline
focal pachymeningitis & 1 & 36 \\ \hline
Glioma & 2 & 76 \\ \hline
Hemorrhagic collection & 1 & 38 \\ \hline
Ischemic change  demyelinating plaque & 1 & 38 \\ \hline
Left Retro-orbital Haemangioma & 2 & 112 \\ \hline
Leukoencephalopathy with subcortical cysts & 1 & 38 \\ \hline
Malformation (Chiari I) & 1 & 38 \\ \hline
Microvascular ischemic change & 2 & 72 \\ \hline
Mid triventricular hydrocephalus & 1 & 38 \\ \hline
NMOSD  ADEM & 1 & 36 \\ \hline
Normal & 44 & 1749 \\ \hline
Obstructive Hydrocephalus & 2 & 76 \\ \hline
Post-operative Status with Small Hemorrhage & 1 & 38 \\ \hline
Postoperative encephalomalacia & 1 & 38 \\ \hline
Small Vessel Diease Demyelination & 1 & 38 \\ \hline
Stroke (Demyelination) & 1 & 38 \\ \hline
Stroke (Haemorrhage) & 13 & 564 \\ \hline
Stroke(infarct) & 19 & 906 \\ \hline
White Matter Disease & 1 & 36 \\ \hline
meningioma & 2 & 76 \\ \hline
pituitary tumor & 1 & 76 \\ \hline
small meningioma & 1 & 36 \\ \hline
\end{tabular}
\caption{National Institute of Neurscience \& Hospitals brain MRI dataset that contains 5285 T1-weighted contrast-enhanced images}
\label{tab:nins_dataset}
\end{table}

\begin{table}[h!]
\tiny
\begin{tabular}{|l|c|l|}
\hline
Category & \multicolumn{1}{l|}{Total Patients} & Total Slices \\ \hline
Meningioma & - & 708 \\ \hline
Glioma & - & 1426 \\ \hline
Pituitary Tumor & - & 930 \\ \hline
\end{tabular}
\caption{Southern Medical University, School of Biomedical Engineering brain MRI dataset that contains 3064 T1-weighted contrast-enhanced images}
\label{tab:biomedical_dataset}
\end{table}

\begin{table}[h!]
\tiny
\begin{tabular}{|l|l|l|}
\hline
Category & Total Patients & Total Slices \\ \hline
Normal & 2 & 65 \\ \hline
Degenerative Disease & 8 & 223 \\ \hline
Neoplastic Disease & 8 & 277 \\ \hline
Inflammatory Infectious Disease & 5 & 189 \\ \hline
Cerebrovascular Disease & 15 & 376 \\ \hline
\end{tabular}
\caption{Harvard Whole Brain Atlas dataset contains 1133 T2-weighted contrast-enhanced images}
\label{tab:harvard_medical_dataset}
\end{table}

\subsection{Essential deep learning techniques} 
In this research, we have applied state-of-the-art techniques in Deep Learning, like, data augmentation, Adam optimizer, Optimal Learning Rate Finder algorithm, and neural network hyper-parameter fine-tuning for the proposed framework.
\subsubsection{Data augmentation:} 
To avoid overfitting the model, which happens because of small training set size, we have used data augmentation techniques such as flipping (vertical and horizontal), rotation, mirroring,  and zooming to create virtual copies of the brain MRI images to improve model generalization performance.
\subsubsection{Optimal Learning Rate Finder (OLRF} 
Hyper-parameter tuning is vital towards model performance improvement, but this process is often cumbersome \cite{smith2017cyclical}.
Therefore, we used the Optimal Learning Rate Finder algorithm to identify a stable set of learning rates to boost model generalization performance. The learning rate indicates the step size used to update the model weights during training. This affects the rate of convergence; if it is too small, convergence to the optimum on the error surface takes much time, and tiny updates on the model weights are performed. Conversely, when the learning rate is too high, the optimization algorithm shoots over the minimum, leading to divergence, thereby affecting model performance. The strategy used to set the step size has a critical role in out of sample generalization performance.
\subsubsection{Stochastic Gradient Descent with Restarts (SGDR)} 
SGDR is a form of learning rate annealing which uses cosine annealing to incrementally reduce the step size during the process of training a neural network. This approach tends to make minimal changes when the optimizer tends towards desired weight parameter updates of the model \cite{huang2017snapshot}.
\subsubsection{Classification Cost function: Softmax Classifier(Multinomial Logistic Cost)}
The goal in a classification task is to maximize log-likelihood or associate higher probability mass for correct class and lower probability mass for incorrect classes. Therefore we define the loss and cost functions are defined as thus:
\begin{dmath*}
	{S =  f(x_i,W)~\text{scores is the unnormalized log}}\\ {\text{probabilities of the classes.}}\\
	{P(Y=k|X=x_i) = \frac{e^{s_k}}{\sum_{j}e^{s_j}}}\\
	{\mathcal{L}_i = -\log P(Y=y_i|X=x_i)}\\
	{\text{Loss Function}~\mathcal{L}_i =  -\log\left(\frac{e^{s_{y_i}}}{\sum_{j}e^{s_j}}\right)}\\
	{\text{Cost Function}~ \mathcal{L}(W) = \frac{1}{N}\sum_{i=1}^N \mathcal{L}_i(f(x_i,W),y_i)~ \{(x_i,y_i)\}_{i=1}^{N}}	
\end{dmath*}

\subsection{Performance Evaluation}\label{experiment}
This subsection presents the metrics we have applied to evaluate the performance of the proposed learning framework. We have used the standard classification performance metric to evaluate the model.
\subsubsection{Performance metrics}
\begin{equation}
    Precision=\frac {A}{\left ({A+\Theta } \right)}
\end{equation}
     
\begin{equation}
    Recall=\frac {A}{\left ({A+B }\right)}
\end{equation}

\begin{equation}
    F1 Score= 2\cdot \frac {\left ({Precision \cdot Recall }\right)}{\left ({Precision +Recall }\right)}
\end{equation}

\begin{equation}
    Accuracy=\frac {A+\Psi}{(P+N)}
\end{equation}

where,

\textbf{False Negatives (B)} are class labels that are predicted negative which in the ground truth are positive.  The is also known as a type two error.

\textbf{False Positives ($\Theta$)} are class examples that are predicted to be positive which in the ground truth are negative.  The is also known as a type one error.

\textbf{True Positive (A)} are class labels that are predicted to be positive which in the ground truth are positive. 

\textbf{True Negative ($\Psi$)} are class labels that are predicted to be negative which in the ground truth are negative. 
\subsection{Simulation Environment} \label{sim_env}
In this research, we carried out all experiments in Python using fast.ai \cite{howard2018fastai}, a wrapper framework built on top of PyTorch library. PyTorch is used for Graphics Processing Unit-based accelerated computing \cite{ketkar2017introduction}. We carried out all experiments on a Linux Server (Ubuntu 10.04.4 LTS) Tesla V100. To evaluate the learning performance, we applied five-fold cross-validation using 80\% - 20\% dataset split.
\section{Experimental Results}
Experimental results of our work throughout the three progressive stages of transfer learning are presented. We present the model performance for the three stages of learning using ResNet50, ResNet34, AlexNet, and VGG19 deep learning CNN architectures. Training the neural network comprises three consecutive steps:
\begin{enumerate}
	\item Training the network with base layers frozen
	\item Retraining the network with augmented data
	\item Unfreezing the Network and fine-tuning with the augmented data
\end{enumerate}

\subsection{Experiment I: National Institute of Neuroscience and Hospitals dataset}\label{experiment:nins}
\subsubsection{Stage I: Retraining the Network}\label{sec:nins_stage_one_analysis}
\begin{figure}[h]
\centering
\includegraphics[width=8.5cm,height=5.4cm]{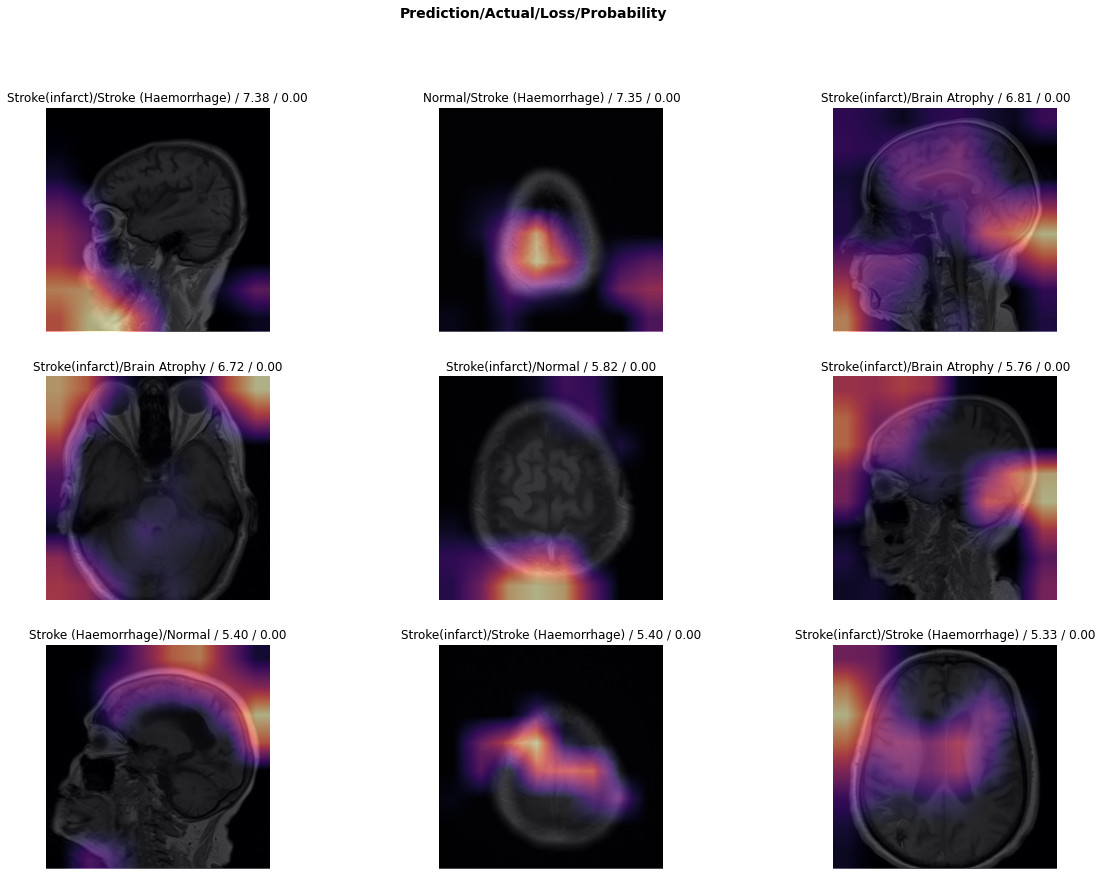}
\caption{Illustration of top miss-classified images after stage I of training}
\label{fig:nins_stage_1}
\end{figure}

In the first stage, we re-trained the model as a benchmark by freezing the convolution base layers of the network and did not update its weights during this stage. The weight updates only occurred in the fully-connected layers of the network. We utilized the multinomial logistic cost function to measure the loss and a step size of $0.01096$. We set the number of epochs to 8 to ensure the model does not overfit on the training set because small, and highly imbalanced dataset tends towards model overfitting, which affects out-of-sample performance. On completing the first stage of the training, the model showed an overall accuracy of 87.43\% using a fivefold cross-validation strategy. Fig \ref{fig:nins_stage_1} shows the top miss-classified images during this phase of training.

We applied OLRF combined with SGDR algorithm to learn a step size in the weight space. This was achieved by boosting the learning rate with respect to the validation loss to discover an optimal learning rate to train the model. SDGR uses cosine annealing to reset the learning rate to traverse regions of the error surface to find the optimal minimum.
\subsubsection{Stage II: Retraining the Network with Augmented Data}\label{sec:nins_stage_two_analysis}
In this second stage, we retrained the model with the augmented data and set the learning rate hyper-parameter by utilizing OLRF and SGDR algorithms between $2e^{-4}$ to $2e^{-2}$.
After this stage of retraining, we did not notice a decrease in the train and validation losses as well as the accuracy as shown in Fig \ref{fig:nins_loss}.

\begin{figure}[h]
\centering
\includegraphics[width=8.5cm,height=5.4cm]{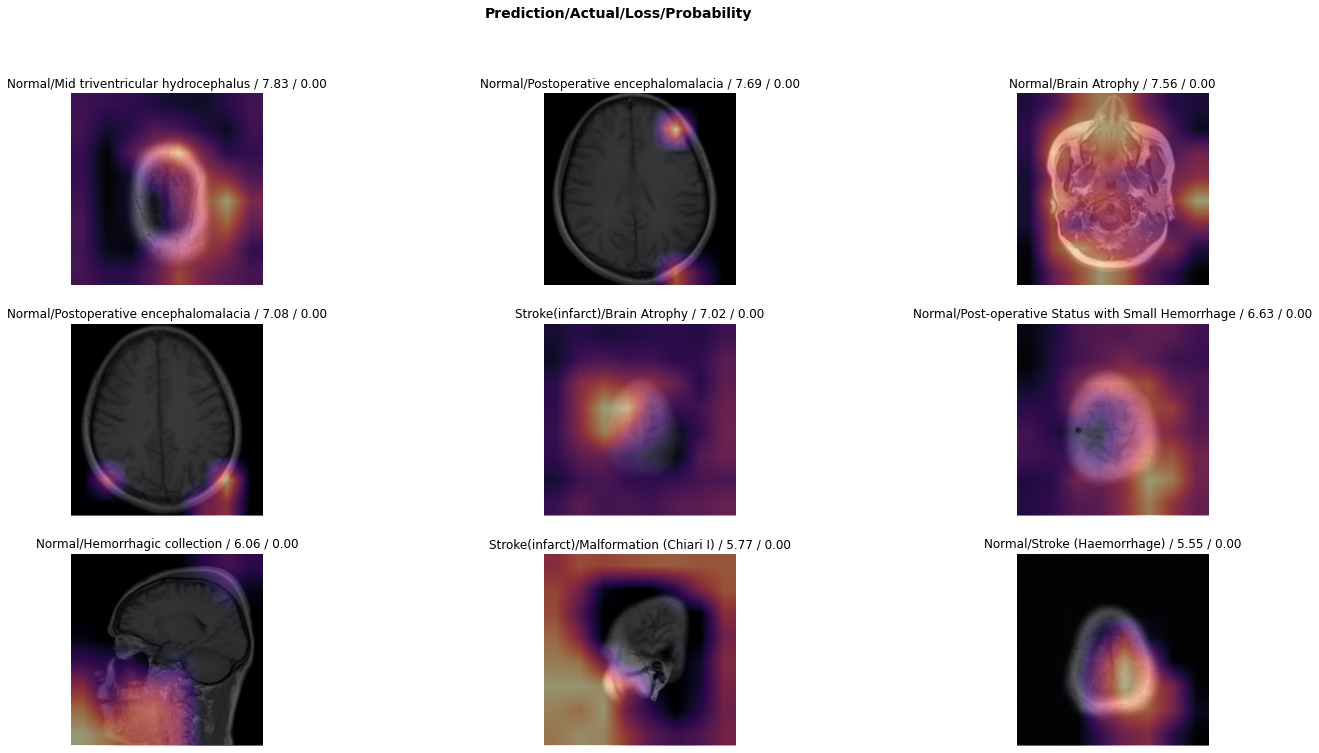}
\caption{Illustration of top miss-classified images after stage II of training with augmented training data}
\label{fig:nins_stage_2_top_loss}
\end{figure}
\subsubsection{Stage III: Unfreezing the Network and Fine-tuning it with the Augmented Data}\label{sec:nins_stage_three_analysis}
\begin{figure}[h]
\centering
\includegraphics[width=8.5cm,height=4.7cm]{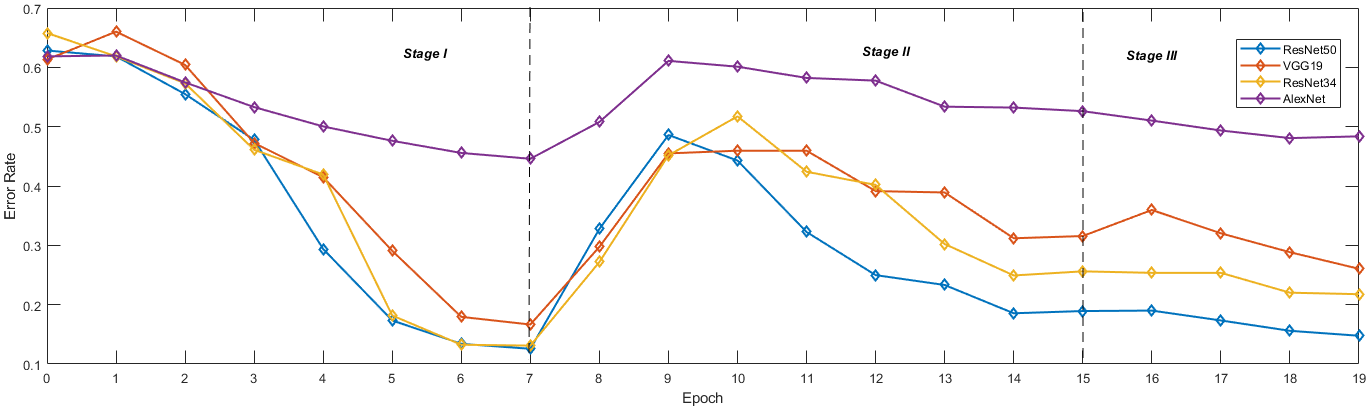}
\caption{Error Rates across the four model architectures}
\label{fig:nins_dataset_error_rates}
\end{figure}

At this third and final stage, we unfroze all the base convolution layers, fine-tuned the network, and jointly trained them using the augmented data. This is Stage-3 in Fig \ref{fig:nins_dataset_error_rates} graph of the error rate across the model architectures. We trained the network for four epochs at this stage to adjust the trained weights to preserve the learned representation. To attain this goal, we set the step size of the last layers higher than the preceding ones during this fine-tuning process. We set the step size in the range of $3e^{-6}$ to $4e^{-3}$ across the network. After fine-tuning, we achieved a validation accuracy of 84.40\% due to the imbalanced nature of the dataset and a degree of noise in it as well.

\begin{figure}[h]
\centering
\includegraphics[width=8.5cm,height=5.4cm]{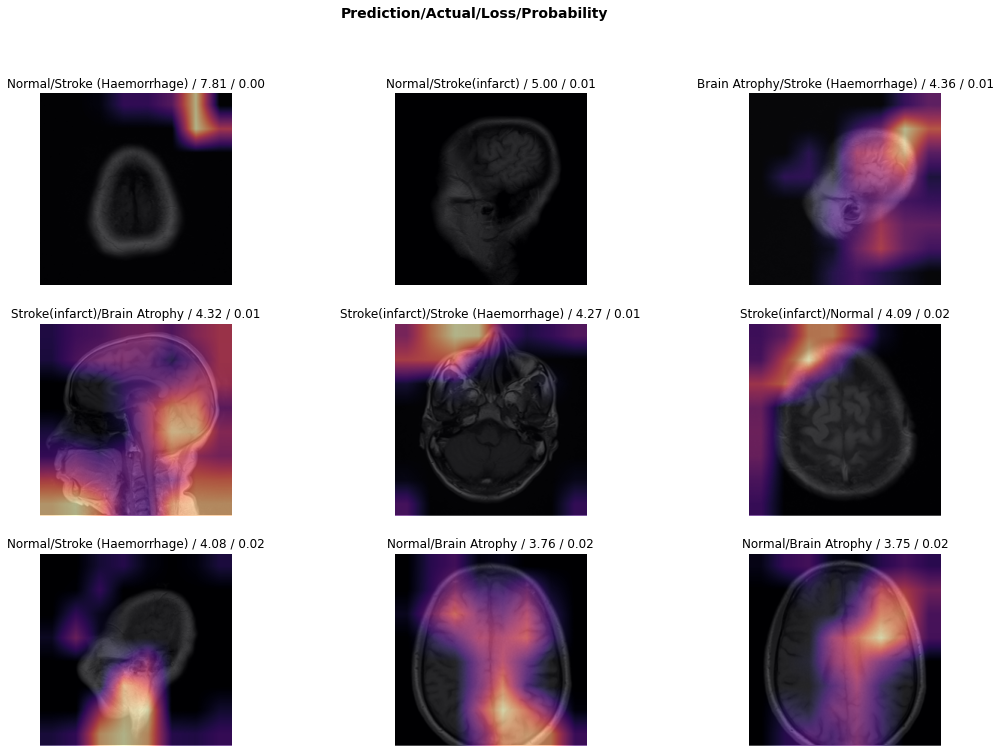}
\caption{Illustration of top miss-classified images after stage III of fine-tuning}
\label{fig:nins_stage_3_top_loss}
\end{figure}
We have plotted the error rates of the models, as shown in Fig.\ref{fig:nins_dataset_error_rates}. ResNet50 converges faster during the last four epochs of the third training stage compared to VGG, Alexnet, and ResNet34. This convergence property of ResNet50 is demonstrated through the nearly smooth learning curve, which indicates the model was able to find a stable set of weights though the data is highly noisy, and classes are imbalanced. The VGG model is second to ResNet50 regarding convergence to the local minimum during the training phases consistently across the three stages of training, which indicates superior performance to Alexnet and ResNet34.

\begin{figure}[h]
\centering
\includegraphics[width=8.5cm,height=4.4cm]{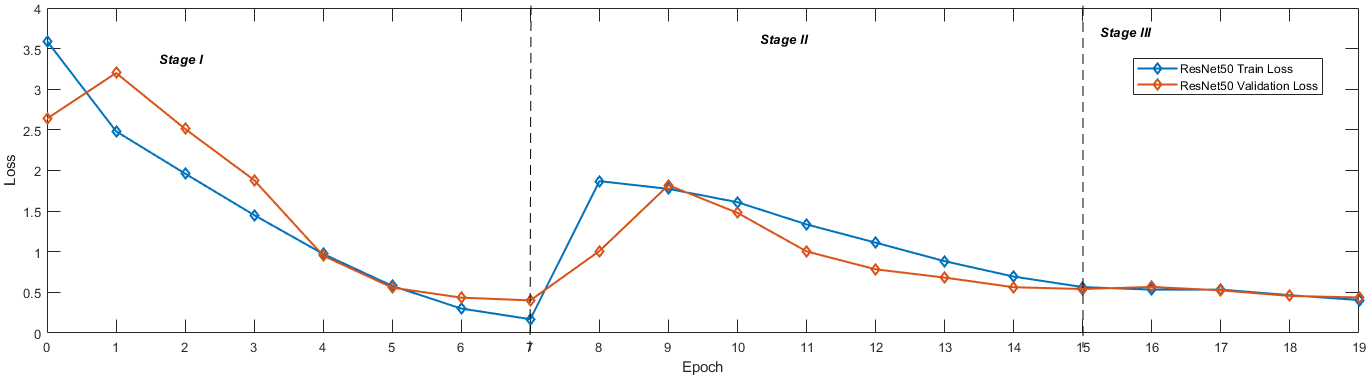}
\caption{Training vs. Validation Loss for all stages with ResNet50}
\label{fig:nins_loss}
\end{figure}

In Table \ref{tab:nins_metrics}, we have presented a comparison of metrics across the three stages of Transfer Learning in this experiment. 
The graph in Fig\ref{fig:nins_loss} gives information on the training and validation accuracy for all training stages for the four CNN model architectures. We observed the network under-fitted since the validation error was lower than the training error. This was solved by extending the number of epochs. At stage two, we used techniques such as vertical flipping, max zooming, and max lighting to augment the MRI images.

\begin{figure}[h]
\centering
\includegraphics[width=8.5cm,height=4.6cm]{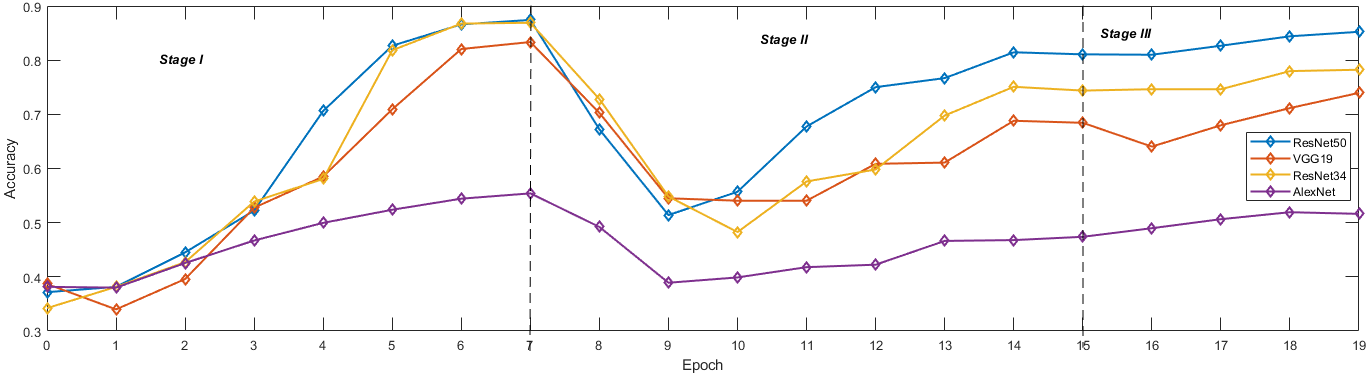}
\caption{Accuracy across four different model architectures}
\label{fig:nins_dataset_accuracy}
\end{figure}

\begin{figure}[h]
\centering
\includegraphics[width=8.5cm,height=4.3cm]{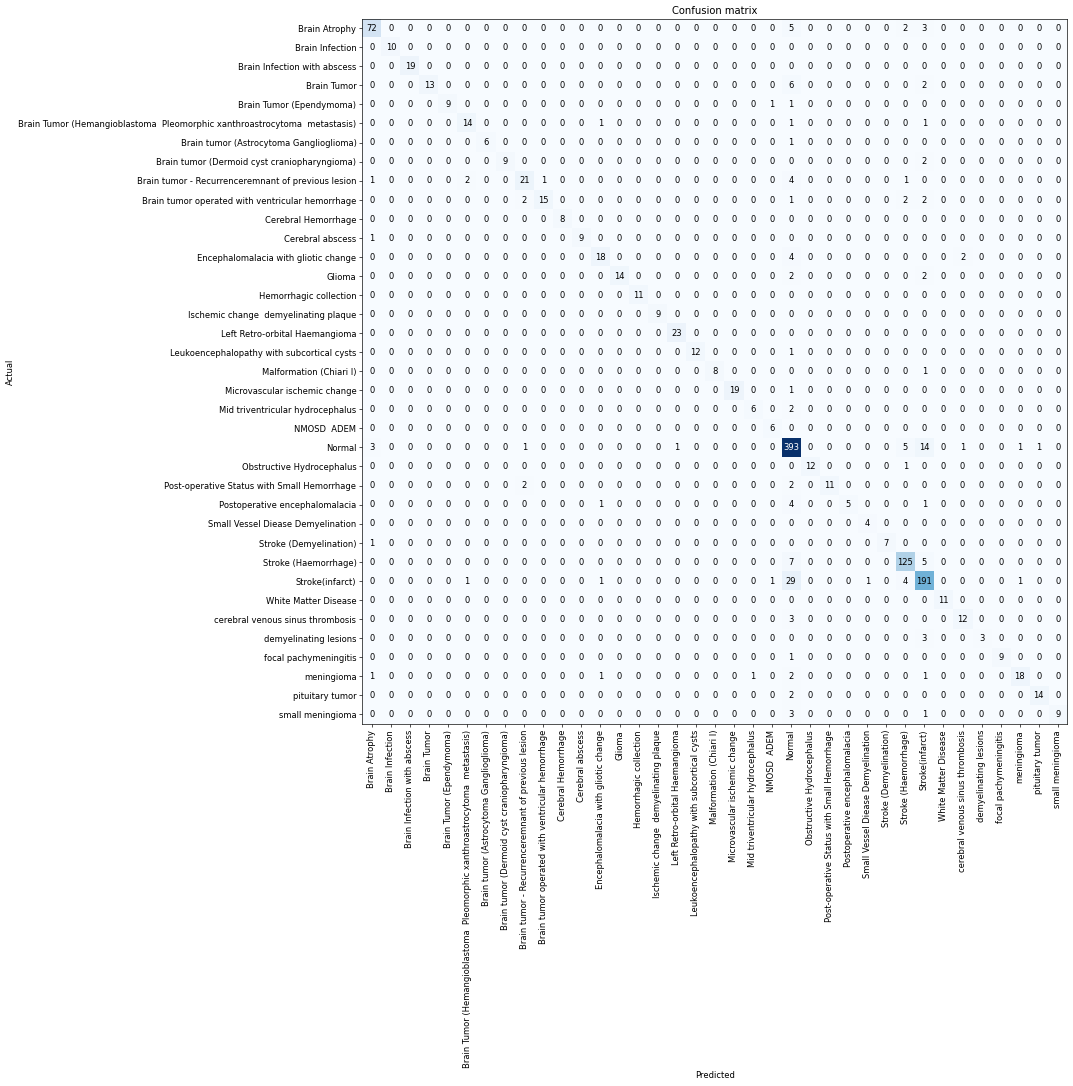}
\caption{Stage I confusion matrix after training the network with base layers frozen}
\label{fig:nins_stage_1_confusion_matrix}
\end{figure}

\begin{figure}[h]
\centering
\includegraphics[width=8.5cm,height=4.3cm]{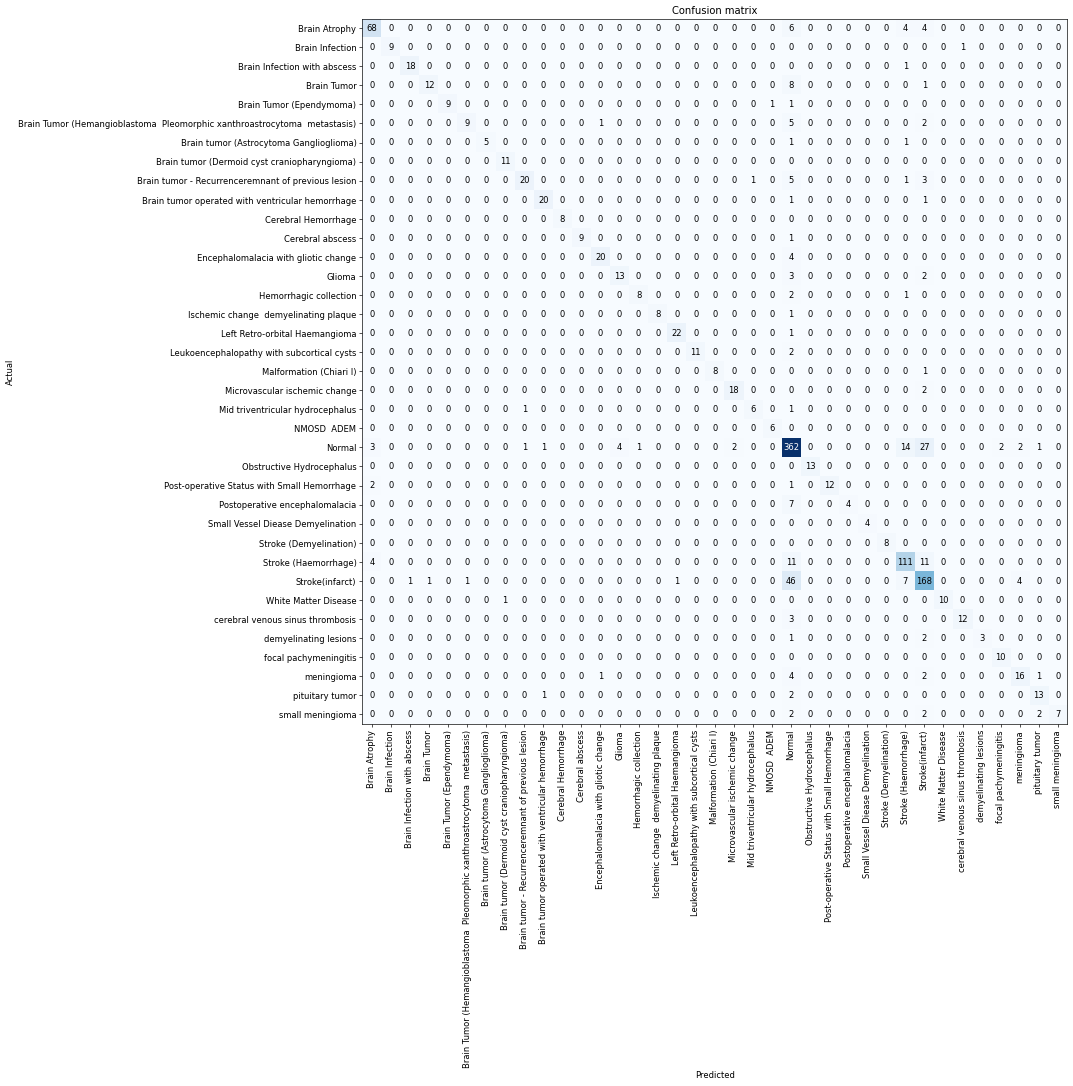}
\caption{Stage II confusion matrix after retraining the network with augmented data}
\label{fig:nins_stage_2_confusion_matrix}
\end{figure}

\begin{figure}[h]
\centering
\includegraphics[width=8.5cm,height=4.5cm]{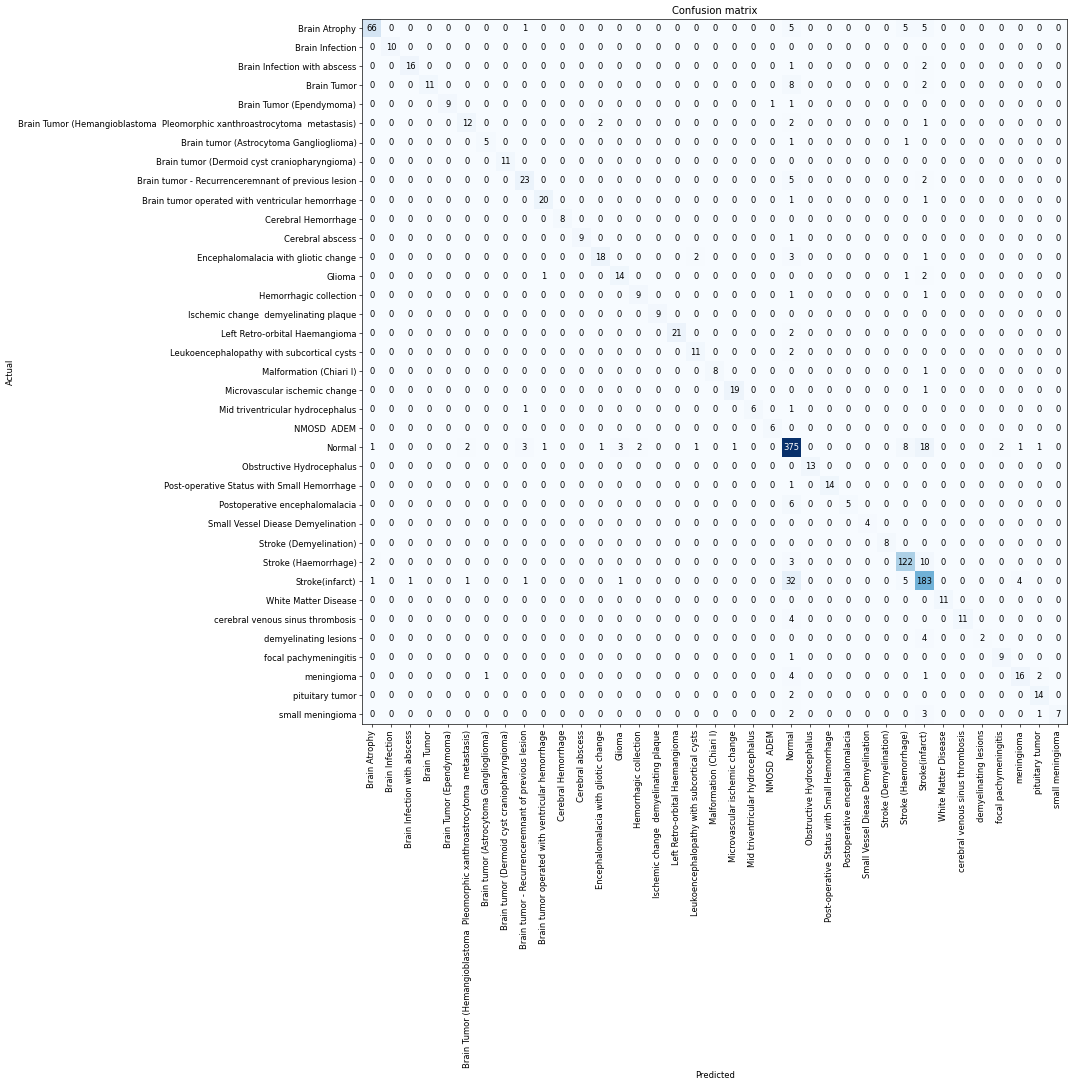}
\caption{Stage III confusion matrix after unfreezing the Network and fine-tuning with the augmented data}
\label{fig:nins_stage_3_confusion_matrix}
\end{figure}

We assess the classification performance on the validation data for the three stages. During the five-fold 
cross-validation training, we obtained the confusion matrices, which are shown 
in Fig.\ref{fig:nins_stage_1_confusion_matrix} through Fig. \ref{fig:nins_stage_3_confusion_matrix}. The matrices give a summary of the classification performance of the model, where the leading diagonals indicate correctly predicted classes and while miss-classified samples are outside the diagonals. At the end of stage I, the model reached an accuracy of 87.43\% as shown in Fig.\ref{fig:nins_dataset_accuracy}. 
However, 
the average five-fold cross-validation accuracy value did not improve at the end of stage 2, which is mostly due to class-imbalance problem with most classes in the dataset having a smaller number of examples.
\begin{table}[]
\tiny
\begin{tabular}{llllllll}
Stage & Train Loss & Validation Loss & Error Rate & Accuracy & Precision & Recall   & F1-Score \\
\hline
I                         &0.169824	&0.401473	&0.125662	&0.874338	&0.874338	&0.874338	&0.874338 \\
II                        &0.565008	&0.541435	&0.189251	&0.810749	&0.810749	&0.810749	&0.810749 \\
III                       &0.434329	&0.460459	&0.155942	&0.844058	&0.844058	&0.844058	&0.844058
\end{tabular}
\caption{Comparison of metrics across the 3 stages of training with NINS dataset}
\label{tab:nins_metrics}
\end{table}

\subsection{Experiment II: Harvard Whole Brain Atlas Dataset}\label{experiment:harvard}
\subsubsection{Stage I: Retraining the Network}\label{sec:harvard_stage_one_analysis}
\begin{figure}[h]
\centering
\includegraphics[width=8.5cm,height=5.4cm]{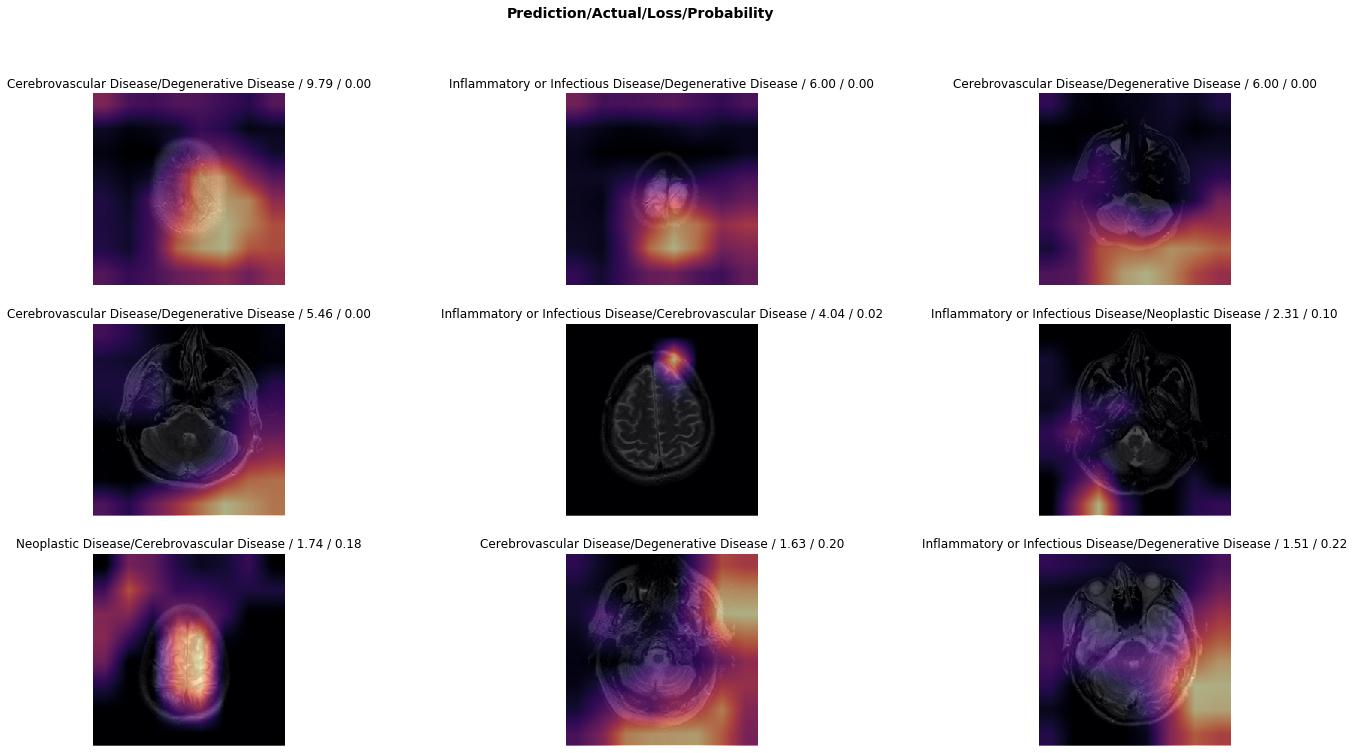}
\caption{Illustration of top miss-classified images after stage I of training}
\label{fig:exp_1_stage_1}
\end{figure}

Following the same methodology used in experiment \ref{experiment:nins}  stage \ref{sec:nins_stage_one_analysis}, we froze the convolution layers of the network and did not update it during this stage. The learning rate hyperparameter was assigned  $1e^{-3}$ and the number of epochs was set to 6. The epoch set to 6 to avoid overfitting the model on the training set. At this first stage, the training model achieved an overall accuracy of  92.47\% in a fivefold cross-validation strategy. Fig \ref{fig:exp_1_stage_1} shows the top miss-classified images during this phase of training.

\subsubsection{Stage II: Retraining the Network with Augmented Data}\label{sec:harvard_stage_two_analysis}
Having completed stage I, we used the OLRF algorithm combined with SDGR to find a stable step size in the weight space. SDGR uses cosine annealing to reset the learning rate to traverse regions of the error surface to find the minimum. We retrained the model with the augmented data and a learning rate set between the intervals $3e^{-3}$ to $2e^{-2}$.
After this stage of retraining, we saw a decrease in the train and validation losses while the accuracy, however, did not increase.
\begin{figure}[h]
\centering
\includegraphics[width=8.5cm,height=5.2cm]{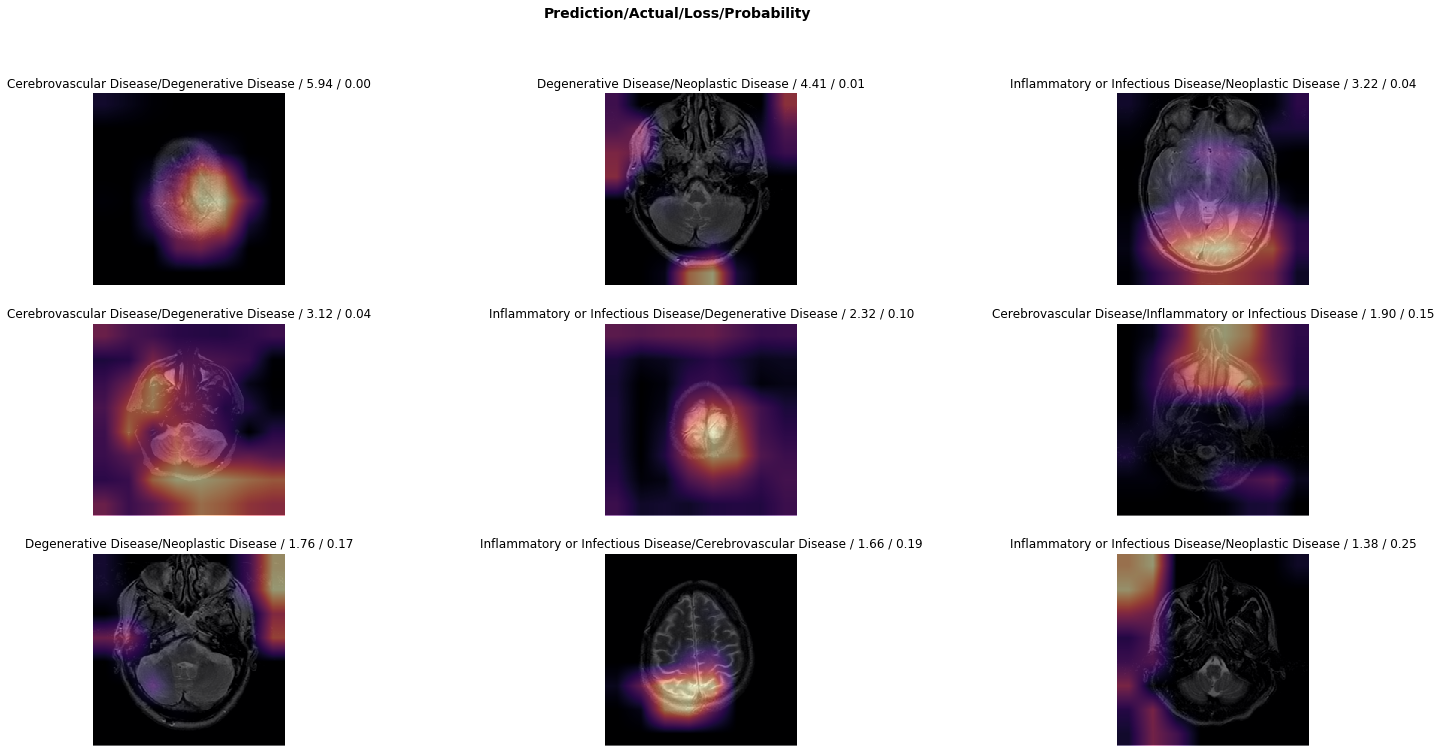}
\caption{Illustration of top miss-classified images after stage II of training with augmented training data}
\label{fig:experiment_2_stage_2_top_loss}
\end{figure}
\subsubsection{Stage III: Unfreezing the Network and Fine-tuning it
with the Augmented Data}\label{sec:harvard_stage_three_analysis}
\begin{figure}[h]
\centering
\includegraphics[width=8.5cm,height=4.8cm]{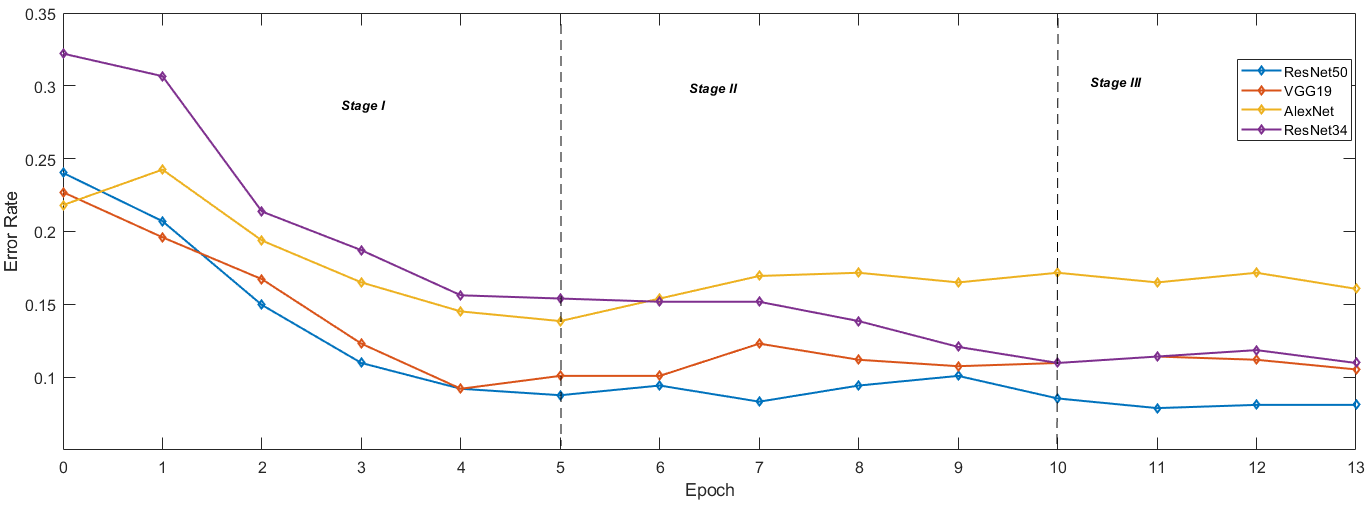}
\caption{Error Rates across the four model architectures}
\label{fig:harvard_dataset_error_rates}
\end{figure}

Similar to the experiment \ref{sec:nins_stage_three_analysis}, we unfroze all the convolution base layers of the network, fine-tuned it, and trained them all using the augmented data. This is the Stage-3 in Fig.\ref{fig:harvard_dataset_error_rates} graph of the error rate across the four model architectures. We have set the step size in the range of $1e^{-6}$ to $4e^{-3}$ across the network. After fine-tuning, we re-trained the model and achieved a validation accuracy of 93.80\%.

\begin{figure}[h]
\centering
\includegraphics[width=8.5cm,height=5.4cm]{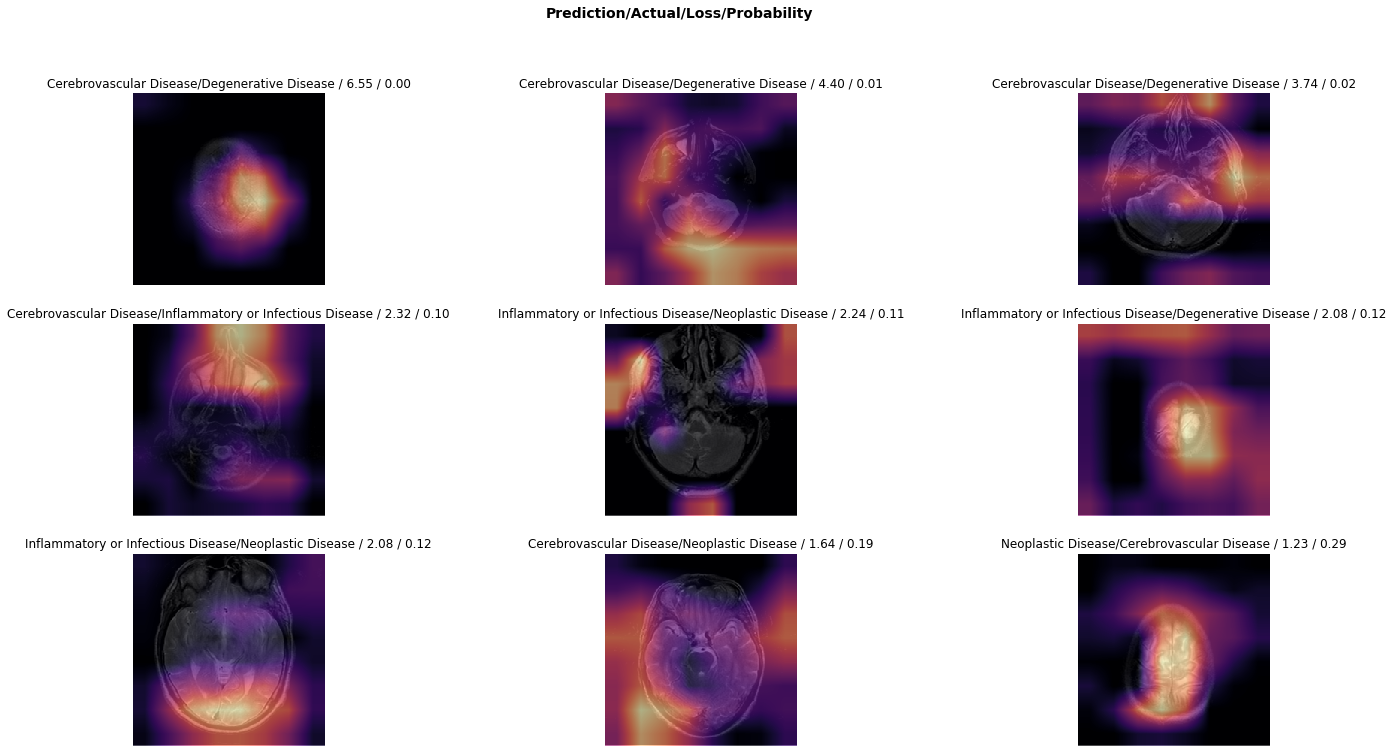}
\caption{Illustration of top miss-classified images after stage III of fine-tuning}
\label{fig:experiment_2_stage_3_top_loss}
\end{figure}
We have plotted the error rates of the models as shown in Fig.\ref{fig:harvard_dataset_error_rates}. ResNet50 converges faster during the last two epochs of the third training stage compared to VGG, Alexnet, and ResNet34. This convergence property of ResNet50 is demonstrated through the nearly smooth learning curve, which indicates the model was able to find a stable set of weights. The VGG model is second to ResNet50 regarding convergence to the local minimum during the training phases consistently across the three stages of training, which indicates superior performance to Alexnet and ResNet34.

\begin{figure}[h]
\centering
\includegraphics[width=8.5cm,height=4.8cm]{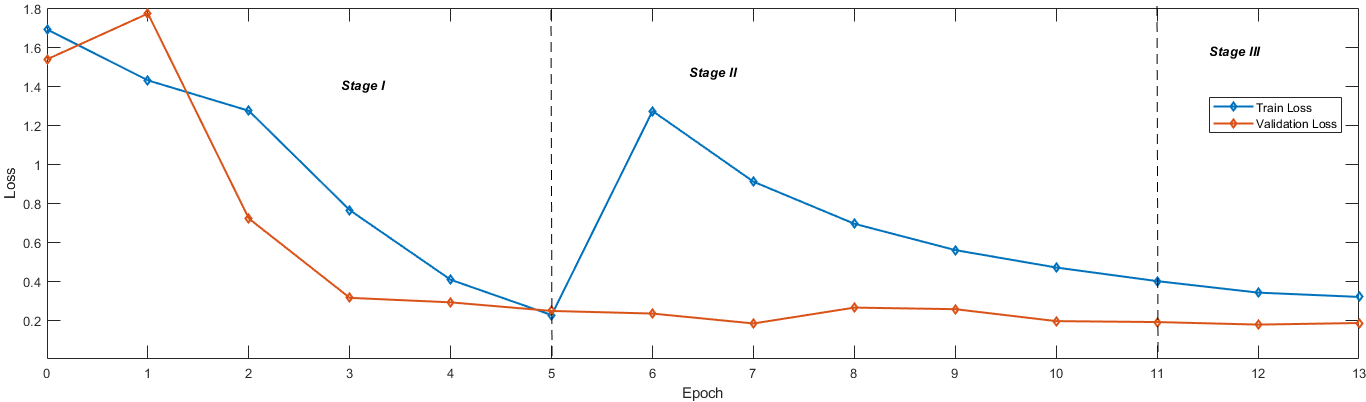}
\caption{Training vs. Validation Loss for all stages with ResNet50}
\label{fig:experiment_2_loss}
\end{figure}

In Table. \ref{tab:expriment_1_metrics}, we have presented the train and validation accuracy graph for the experiment. 
The graph in Fig.\ref{fig:experiment_2_loss} gives information on the training and validation accuracy for all three stages of training. We observed the network under-fitted since the validation error was lower than the training error. This was solved by extending the number of epochs. At stage two, we used techniques such as vertical flipping, max zooming, and max lighting to augment the MRI images.

\begin{figure}[h]
\centering
\includegraphics[width=8.5cm,height=4.3cm]{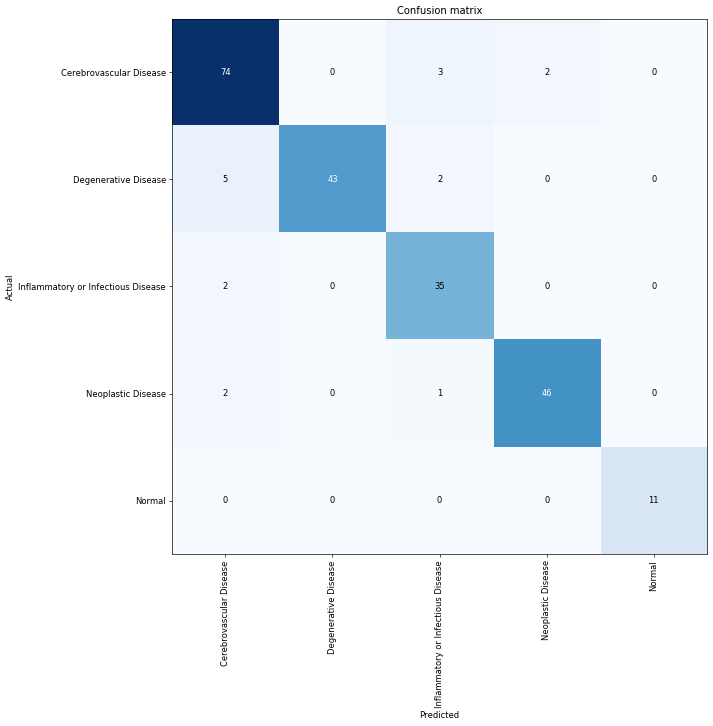}
\caption{Stage I confusion matrix after training the network with base layers frozen}
\label{fig:experiment_2_stage_1_confusion_matrix}
\end{figure}

\begin{figure}[h]
\centering
\includegraphics[width=8.5cm,height=4.3cm]{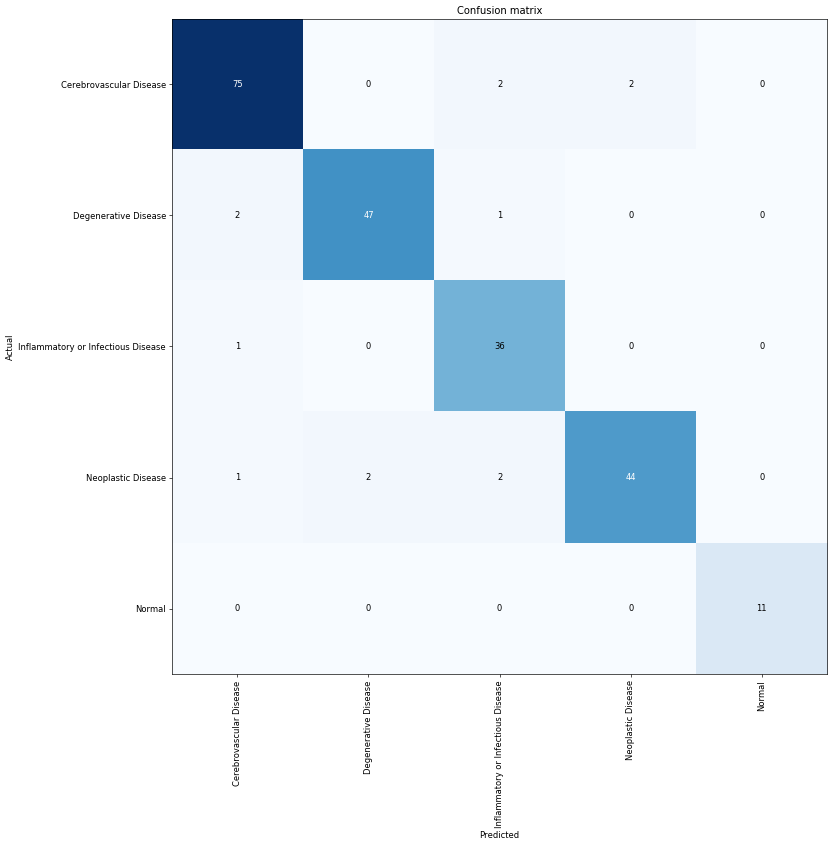}
\caption{Stage II confusion matrix after retraining the network with augmented data}
\label{fig:experiment_2_stage_2_confusion_matrix}
\end{figure}

\begin{figure}[h]
\centering
\includegraphics[width=8.5cm,height=4.5cm]{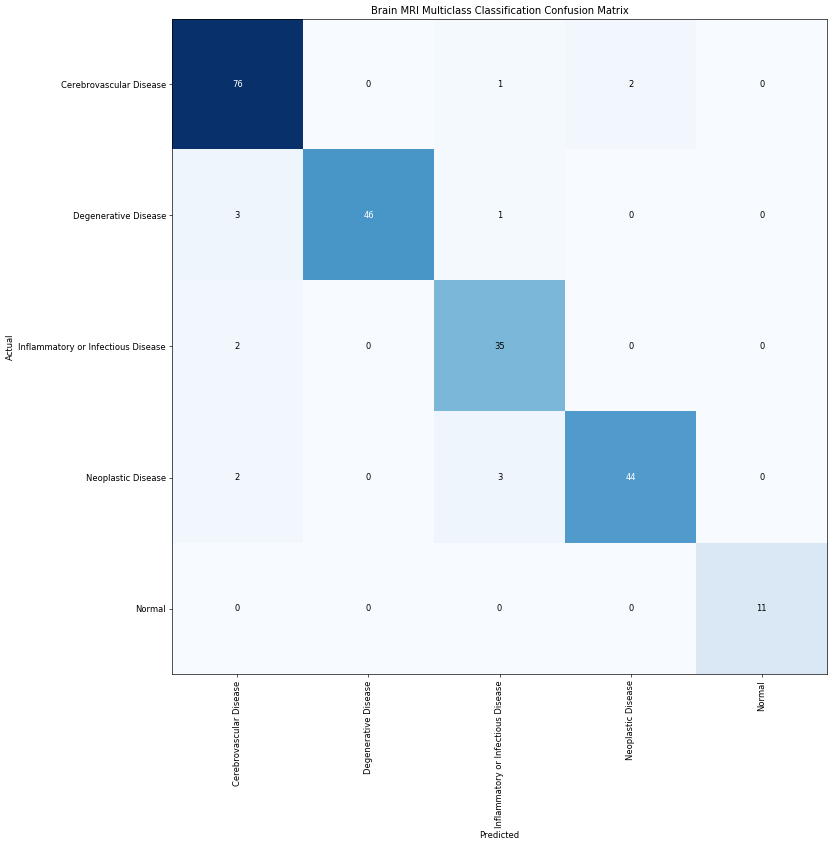}
\caption{Stage III confusion matrix after unfreezing the Network and fine-tuning with the augmented data}
\label{fig:experiment_2_stage_3_confusion_matrix}
\end{figure}

We assess the classification performance on the validation data for the three stages. During the five-fold 
cross-validation training, we obtained the confusion matrices, which are presented 
in Fig.\ref{fig:experiment_2_stage_1_confusion_matrix} through Fig. \ref{fig:experiment_2_stage_3_confusion_matrix}. The matrices give a summary of the classification performance of the model were the leading diagonals indicate correctly predicted classes and while miss-classified samples are outside the diagonals. At the end of stage I, the model reached an accuracy of 92.47\% with 17 incorrectly classified images, as shown in Fig.\ref{fig:experiment_2_stage_1_confusion_matrix}. This model correctly classified all normal cases in stage II, as shown in  Fig. \ref{fig:experiment_2_loss}.

\begin{figure}[h]
\centering
\includegraphics[width=8.5cm,height=4.8cm]{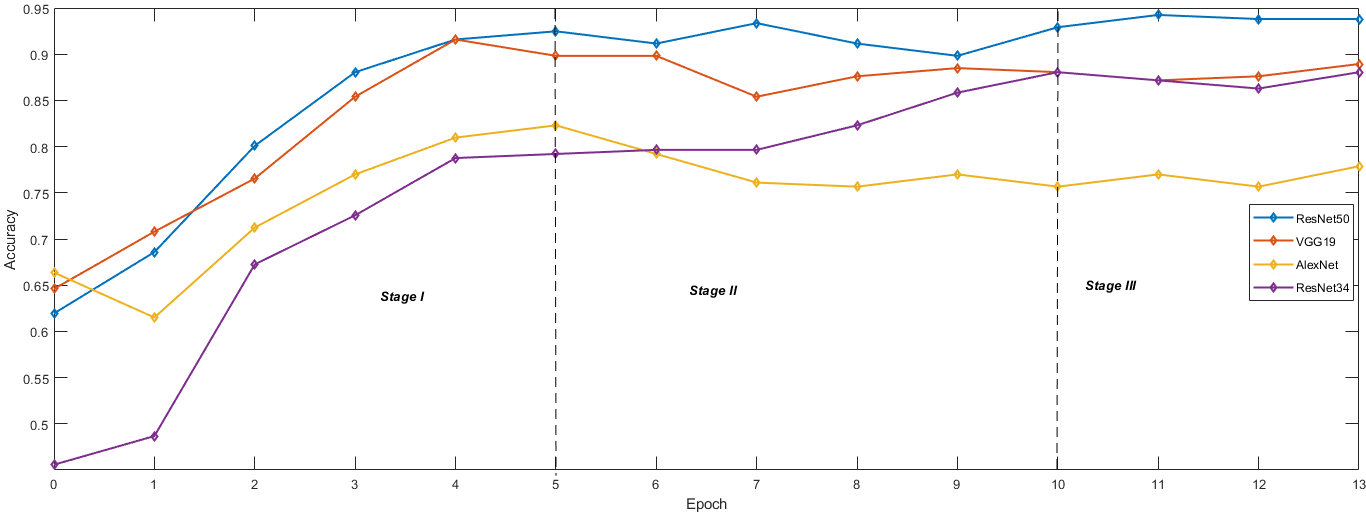}
\caption{Accuracy across four different model architectures}
\label{fig:harvard_dataset_accuracy}
\end{figure}
However, the average five-fold cross-validation accuracy value improved from 92.24\% to 94.42\% at the end of stage 2. 
Finally, Fig.\ref{fig:experiment_2_stage_3_confusion_matrix} depicts  stage III results. 14 images were incorrectly classified, and the model reached 93.80\% average five-fold cross-validation accuracy. The whole training procedure 
takes 310 seconds.
\begin{table}[]
\tiny
\begin{tabular}{llllllll}
Stage & Train Loss & Validation Loss & Error Rate & Accuracy & Precision & Recall   & F1-Score \\
\hline
I                         & 0.225288   & 0.246468        & 0.075221   & 0.924779 & 0.924779  & 0.924779 & 0.924779 \\
II                        & 0.399370   & 0.189434        & 0.057522   & 0.942478 & 0.942478  & 0.942478 & 0.942478 \\
III                       & 0.318317   & 0.184569        & 0.061947   & 0.938053 & 0.938053  & 0.938053 & 0.938053
\end{tabular}
\caption{Comparison of metrics across the 3 stages of training with Harvard dataset}
\label{tab:expriment_1_metrics}
\end{table}

\subsection{Experiment III: Biomedical School Brain MRI dataset}
\subsubsection{Stage I: Retraining the Network}\label{sec:biomedical_stage_one_analysis}

\begin{figure}[h]
\centering
\includegraphics[width=8.5cm,height=8.5cm]{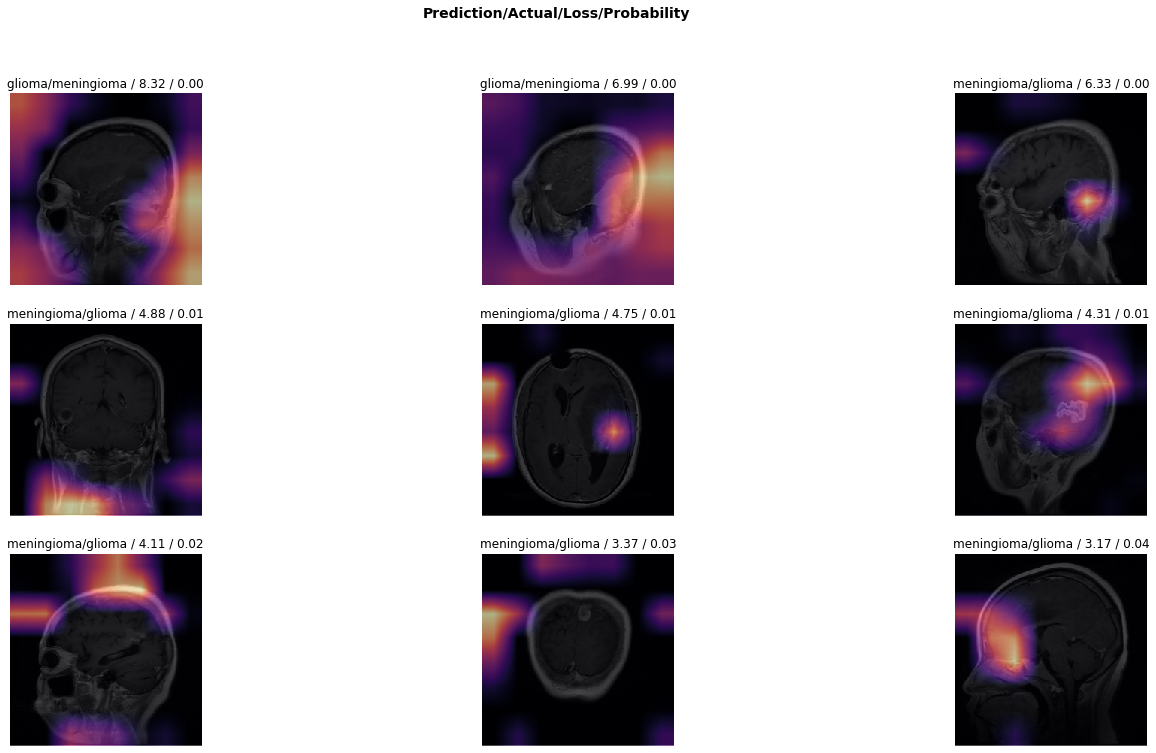}
\caption{ Illustration of top miss-classified images after stage I of training}
\label{fig:experiment_1_stage_1_top_loss}
\end{figure}

In this third and final experiment, we used the brain MRI dataset from the School of Biomedical Engineering, which consists of T-1 weighted images. We re-trained the model as a benchmark. We froze the convolution layers of the network and did not update it during this stage. We only trained the weights of the fully-connected layers of the model. We used the multinomial logistic cost function to measure the loss and error rate with a step size of $1e^{-3}$ and trained the model for 4 epochs. We set the number of epochs to 4 to ensure the model does not overfit on the training set. The model memorises the given small dataset which affects out-of-sample performance. On completing the first stage of the training, the model showed an overall accuracy of 96.73\% using a fivefold cross-validation strategy. Fig \ref{fig:experiment_1_stage_1_top_loss} shows the top miss-classified images during this phase of training.

\subsubsection{Stage II: Retraining the Network with Augmented Data}\label{sec:biomedical_stage_two_analysis}
Similar to the preceding experiments, we utilized the OLRF with SGDR algorithms to learn an optimal step size in the weight space. This is achieved by boosting the learning rate with respect to the validation loss to discover an optimal learning rate to train the model. We used an initial learning rate set between the intervals $3e^{-3}$ to $2e^{-2}$  in this stage of the experiment. SDGR uses cosine annealing to reset the learning rate to traverse regions of the error surface to find the minimum.
After this stage of retraining, we saw a decrease in the train and validation losses, as shown in Fig.\ref{fig:experiment_1_loss}. We used techniques such as vertical flipping, max zooming, and max lighting to augment the MRI images. The performance metrics are presented in Table. \ref{tab:expriment_2_metrics}.

\begin{figure}[h]
\centering
\includegraphics[width=8.5cm,height=6.6cm]{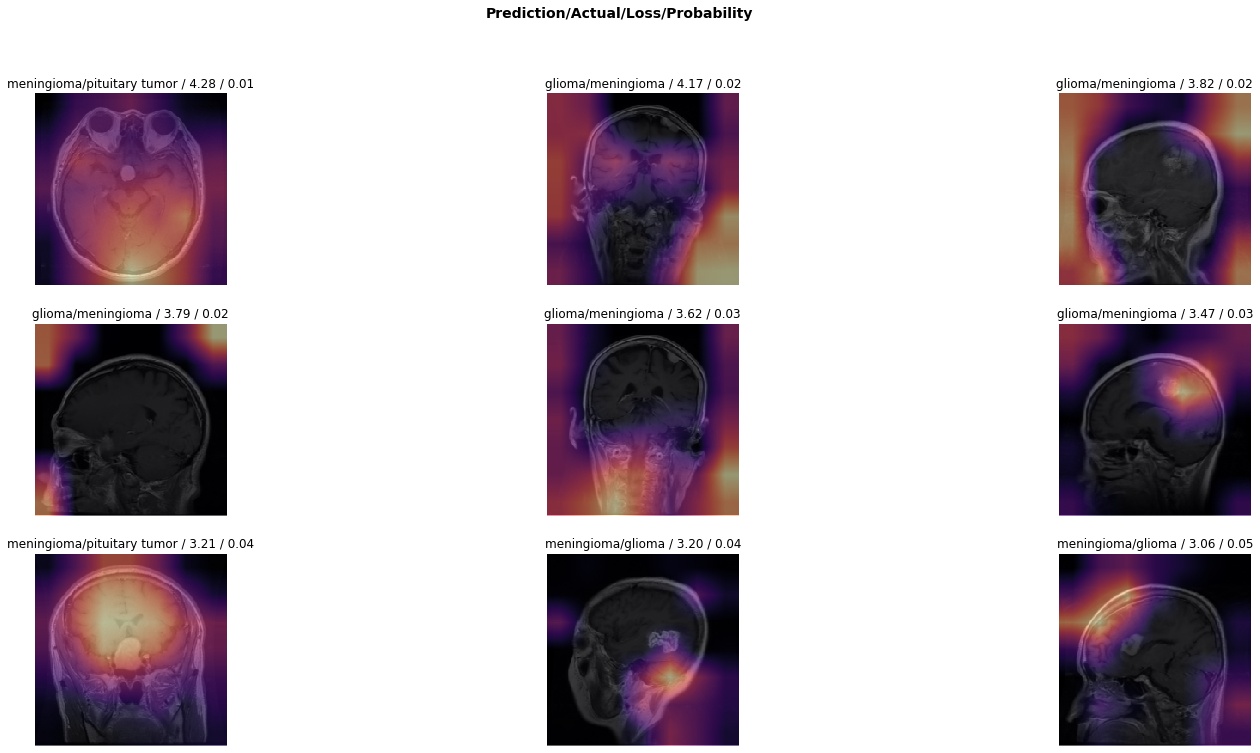}
\caption{Illustration of top miss-classified images after stage II of training with augmented training data}
\label{fig:experiment_1_stage_2_top_loss}
\end{figure}
\subsubsection{Stage III: Unfreezing  the  Network  a and fine-tuning it with the Augmented Data}\label{sec:biomedical_stage_three_analysis}
\begin{figure}[h]
\includegraphics[width=8.5cm,height=4.9cm]{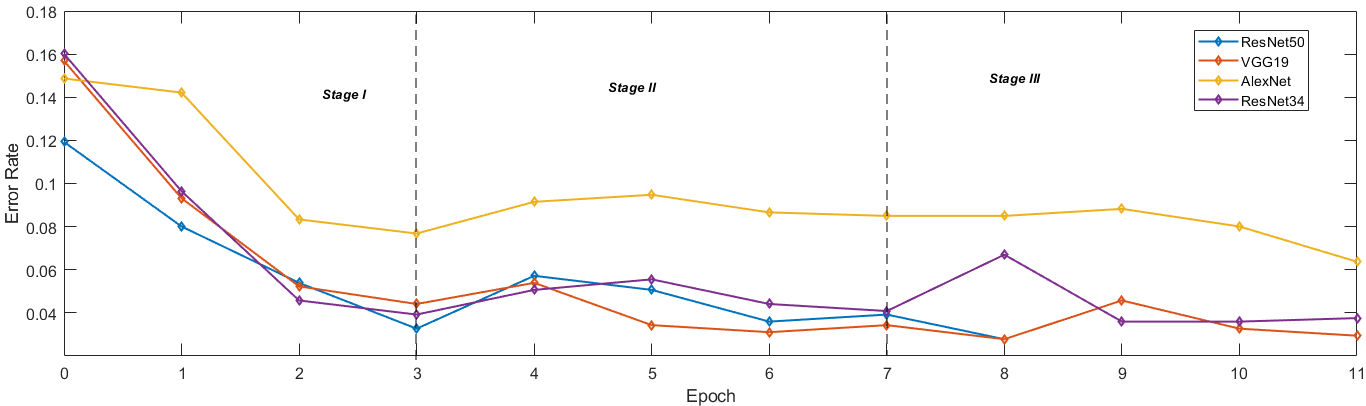}
\caption{Error Rates across four different model architectures}
\label{fig:biomedical_dataset_error_rates}
\end{figure}

Similar to \ref{sec:nins_stage_three_analysis} and \ref{sec:harvard_stage_three_analysis}, we unfroze all the convolution layers of the network, fine-tuned the it, and jointly trained them using the augmented data. This is the Stage-3 in Fig.\ref{fig:biomedical_dataset_error_rates} graph of the error rate across the model architectures. At this stage, we trained the fully connected layers for 4 epochs, and we want the trained weights to adjust in line with training steps. To attain this goal, we set the step size of the last layers higher than the preceding layers during this process of fine-tuning. For this reason, we varied the learning rate across the layers of the network. At this stage, we set the learning rate to $1e^{-6}$ through $4e^{-3}$ across the network. After fine-tuning, we re-trained the model and achieved a validation accuracy of 97.05\%.

\begin{figure}[h]
\centering
\includegraphics[width=8.5cm,height=6.6cm]{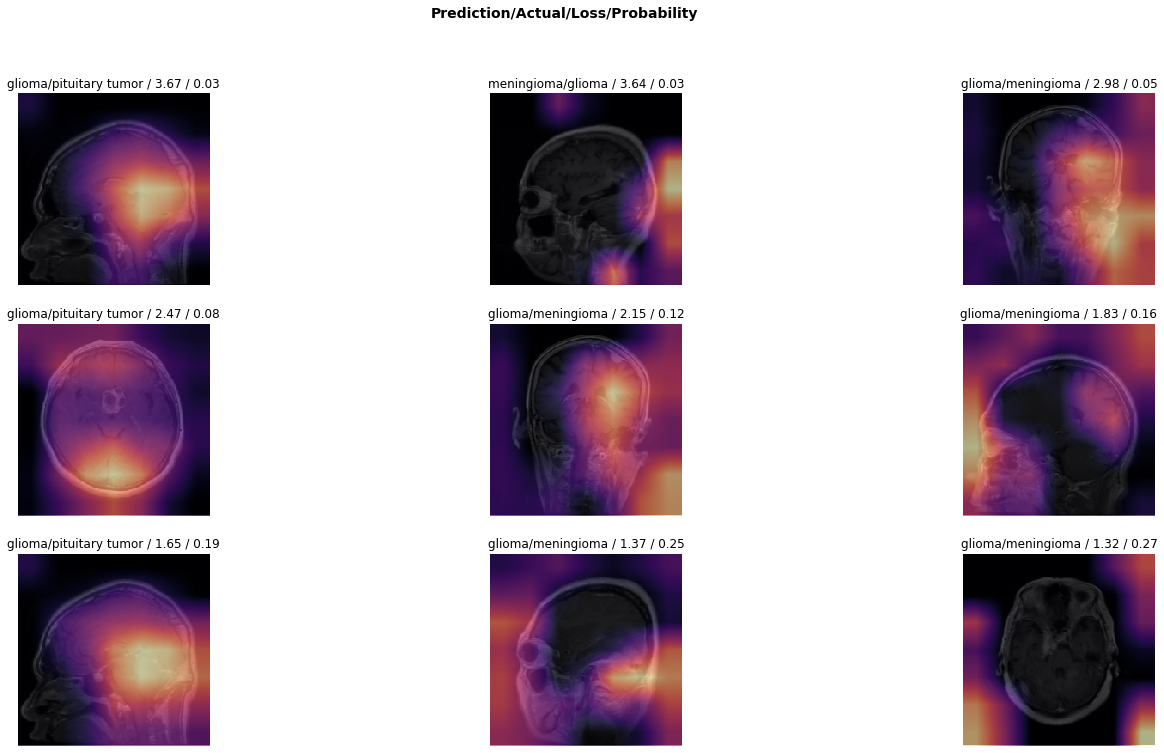}
\caption{Illustration of top miss-classified images after stage III of fine-tuning}
\label{fig:experiment_1_stage_3_top_loss}
\end{figure}

\begin{figure}[h]
\includegraphics[width=8.5cm,height=4.9cm]{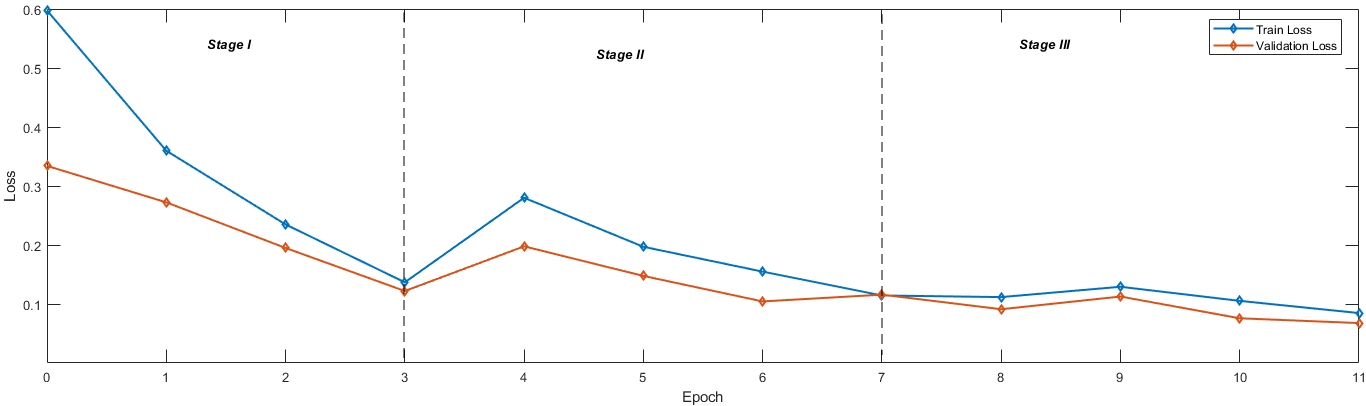}
\caption{Training vs. Validation Loss for all stages with ResNet50}
\label{fig:experiment_1_loss}
\end{figure}

We have plotted the error rates of the models, as presented in Fig.\ref{fig:biomedical_dataset_error_rates}. ResNet50 converges faster during the last two epochs of the third training stage compared to VGG, Alexnet, and ResNet34. This convergence property of ResNet50 is demonstrated through the nearly smooth learning curve, which indicates the model was able to find a stable set of weights. The VGG model is second to ResNet50 regarding convergence to the local minimum during the training phases consistently across the three stages of training, which indicates superior performance to Alexnet and ResNet34.
In Table. \ref{tab:expriment_2_metrics}, we have presented the training and validation accuracy values for the experiment. 
The graph in Fig.\ref{fig:experiment_1_loss} gives information on the training and validation accuracy for all stages of training. We observed the network under-fitted since the validation error was lower than the training error. This was solved by extending the number of epochs. At stage two, we used techniques such as vertical flipping, max zooming, and max lighting to augment the MRI images.

\begin{figure}[h]
\centering
\includegraphics[width=8.5cm,height=4.4cm]{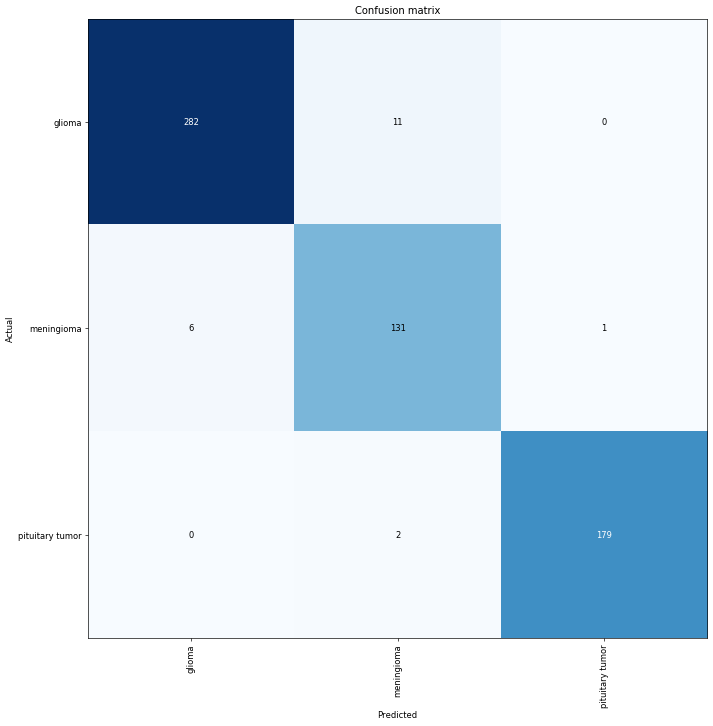}
\caption{Stage I confusion matrix after training the network with base layers frozen}
\label{fig:experiment_1_stage_1_confusion_matrix}
\end{figure}

\begin{figure}[h]
\centering
\includegraphics[width=8.5cm,height=4.4cm]{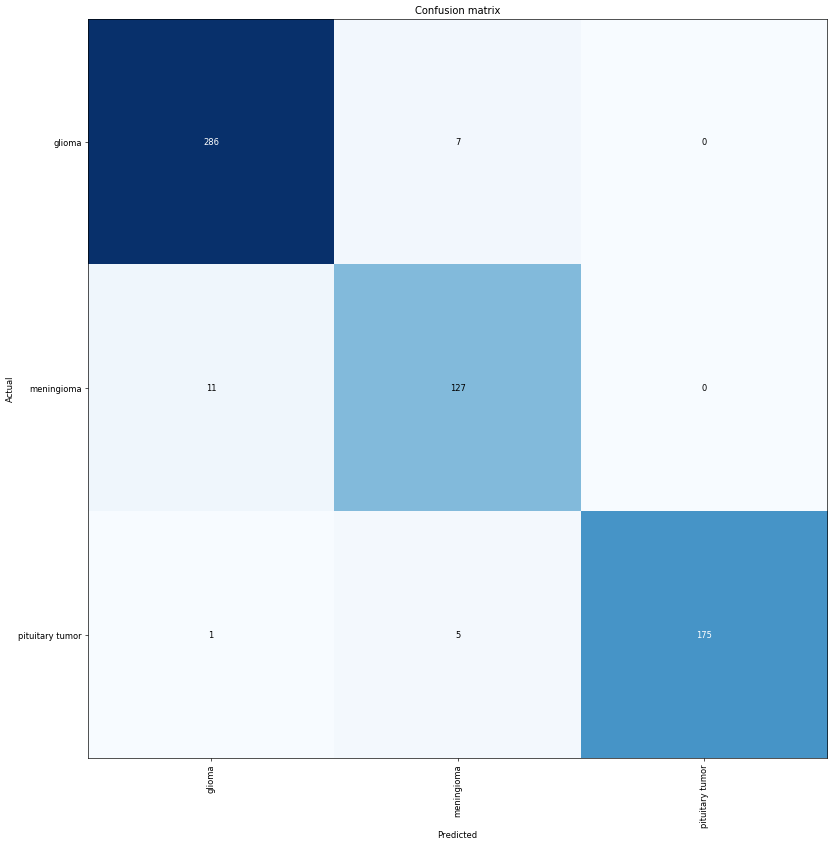}
\caption{ Stage II  confusion matrix after retraining the net-work with augmented data}
\label{fig:experiment_1_stage_2_confusion_matrix}
\end{figure}

\begin{figure}[h]
\centering
\includegraphics[width=8.5cm,height=4.4cm]{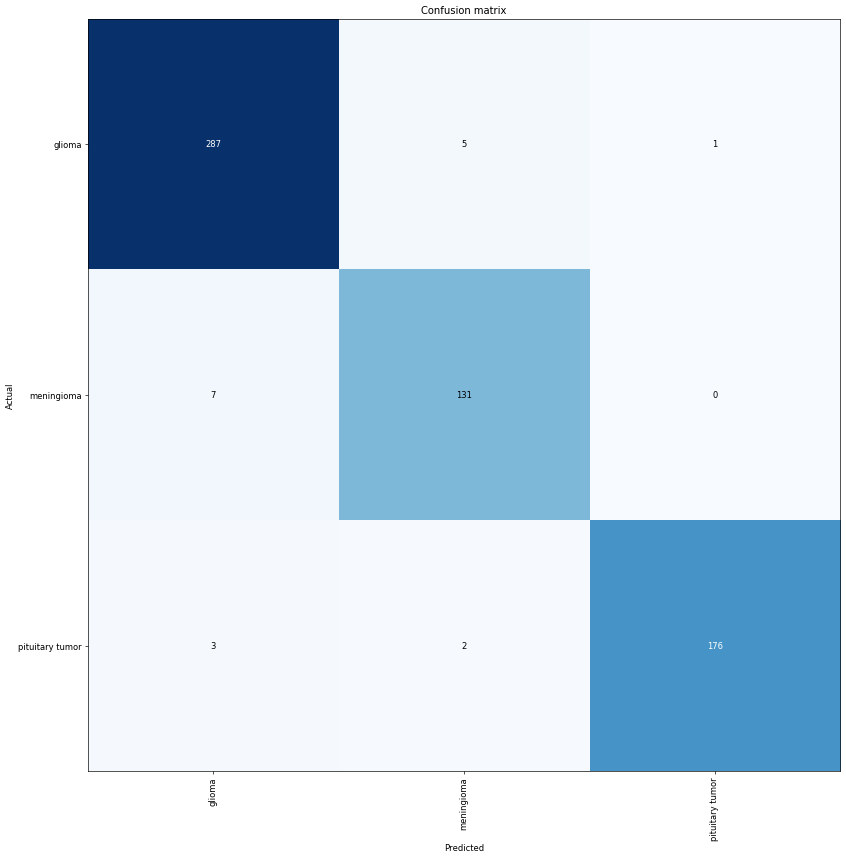}
\caption{Stage III confusion  matrix  after  unfreezing  the Network and fine-tuning with the augmented data}
\label{fig:experiment_1_stage_3_confusion_matrix}
\end{figure}

\begin{table}[]
\tiny
\begin{tabular}{llllllll}
Stage & Train Loss & Validation Loss & Error Rate & Accuracy & Precision & Recall   & F1-Score \\
\hline
I                         & 0.136737   & 0.121995        & 0.032680   & 0.967320 & 0.967320  & 0.967320 & 0.967320 \\
II                        & 0.114440   & 0.104330        & 0.035948   & 0.964052 & 0.964052  & 0.964052 & 0.964052 \\
III                       & 0.084613   & 0.067653        & 0.029412   & 0.970588 & 0.970588  & 0.970588 & 0.970588
\end{tabular}
\caption{Comparison of metrics across the 3 stages of training with School of Biomedical Engineering dataset}
\label{tab:expriment_2_metrics}
\end{table}
We assess the classification performance on the validation data for the three stages. During the five-fold 
cross-validation training, we obtained the confusion matrices, which are shown 
in Fig.\ref{fig:experiment_1_stage_1_confusion_matrix} through Fig. \ref{fig:experiment_1_stage_3_confusion_matrix}. The matrices give a summary of the classification performance of the model were the leading diagonals indicate correctly predicted classes and while miss-classified samples are outside the diagonals. At the end of stage I, the model reached an accuracy of 96.73\% with 20 incorrectly classified images across the three classes, as shown in Fig.\ref{fig:experiment_1_stage_1_confusion_matrix}. The model in stage II, as shown in  Fig. \ref{fig:experiment_2_loss} had a steady number of incorrectly classified images.
\begin{figure}[h]
\centering
\includegraphics[width=8.5cm,height=4.9cm]{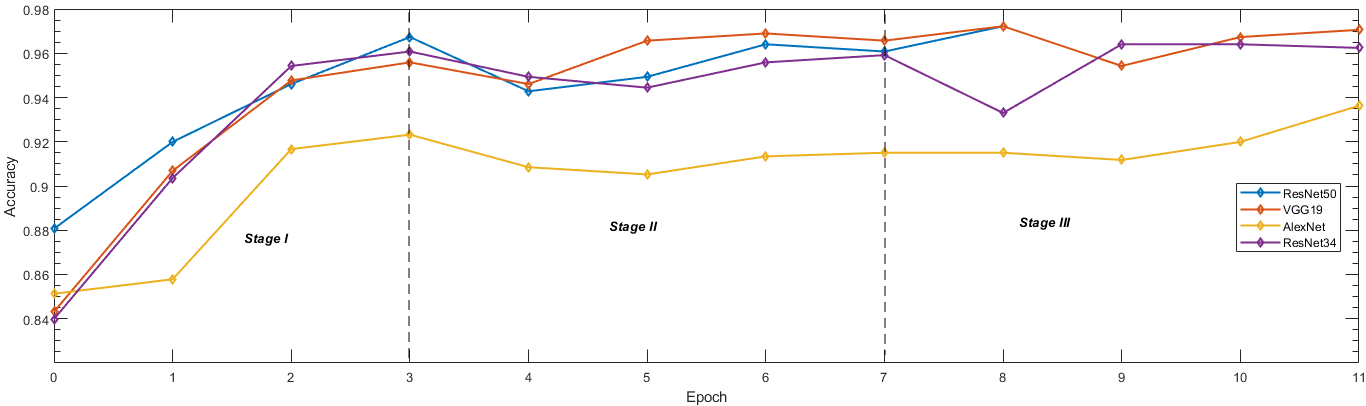}
\caption{Accuracy across four different model architectures}
\label{fig:biomedical_dataset_accuracy}
\end{figure}
However, 
the average five-fold cross-validation accuracy value improved from 96.73\%  to  97.05\% at the end of stage III. 
Finally, Fig.\ref{fig:experiment_1_stage_3_confusion_matrix} depicts  stage III results. 17 images were incorrectly classified and the model reached 97.05\% average five-fold cross-validation accuracy. The whole training procedure 
takes 430 seconds.

\section{Discussion and Conclusion}
\subsection{Discussion}
\begin{figure}[h]
\centering
\includegraphics[width=8.5cm,height=5.3cm]{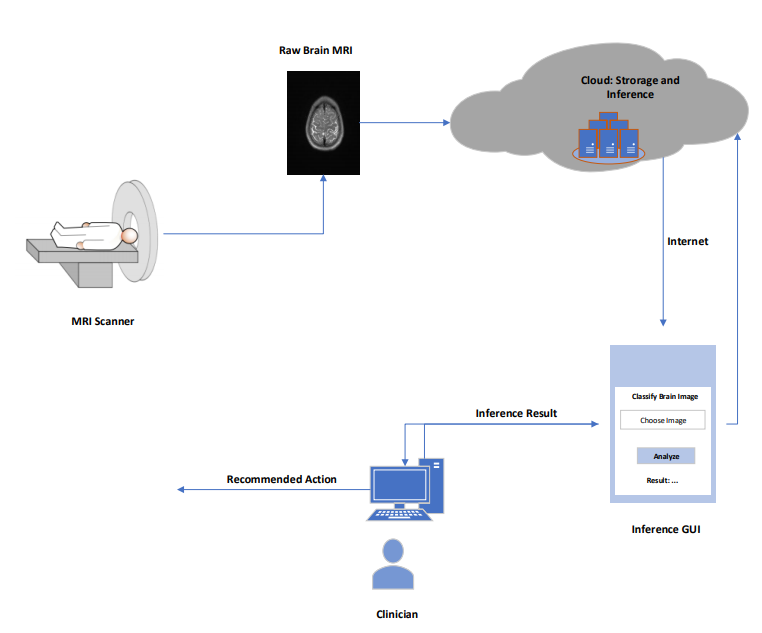}
\caption{Proposed system deployment diagram}
\label{fig:system_deployment}
\end{figure}

Many researchers have proposed methods and techniques in Machine Learning using standard publicly available datasets for brain MRI classification tasks. These methods vary in their approaches from feature engineering, dimensionality reduction to deep learning-based. Table \ref{tab:research_comparison} presents a comparison of research on the classification of MRI brain images. The table presents our proposed dataset Study I and three publicly available datasets:  Study II (Harvard Whole Brain Atlas) \cite{Summers288}, Study III (The Cancer Imaging Archive) \cite{clark2013cancer}, Study IV (School of Biomedical Engineering, Guangzhou University) \cite{cheng2015correction}. 
In their paper, Chaplot et al. \cite{chaplot2006classification} proposed a 3-level Wavelet Transform method for feature extraction and a Support Vector Machine binary brain MRI classifier using 52 MRI datapoints obtaining an accuracy of 98\%. Gudigar et al. \cite{gudigar2019application} in a comparative study assessed three multi-resolution analysis methods - Discrete Wavelet, Curvelet, and Shearlet Transform, and Particle Swarm Optimization for textual feature extraction from transformed images using SVM classifier reported accuracy of 97.38\%  with a dataset of 612 brain MRI scans based on Shearlet Transform. In their 2015 paper, Zhang et al. \cite{zhang2015preclinical},  proposed a novel Computer-Aided Diagnostic system that comprises Discrete Wavelet Packet Transform (DWPT), which serves as a Wavelet Packet Coefficient Extractor, Shannon and Tsallis entropies which obtained entropy features from DWPT. Finally, they used a Generalized Eigenvalue Proximate Support Vector Machine with Radial Basis Function for brain MRI binary classification achieved 99.33\% accuracy using a dataset of 255 brain MRI images. Saritha et al. \cite{saritha2013classification} have used Wavelet Entropy-based Spider Web Plots' for dimensionality reduction combined with a Probabilistic Neural Network (PNN) for the classification reported accuracy of 100\%. El-Dahshan et al. \cite{el2014computer} in their work presented a novel technique consisting of  Discrete Wavelet Transform, PCA to reduce the feature vectors dimensions, and a Feed-Forward Artificial Neural Network to classify 101 brain MRI images achieved 98.6\% accuracy. Wang et al. \cite{wang2015feed} used a pipeline of Stationary Wavelet Transform for feature extraction, Principal Component Analysis to reduce the feature space combined with Particle Swarm Optimization and Artificial Vee Colony to classify MRI brain images reported accuracy of 99.45\%. Nayak et al.\cite{nayak2016brain} proposed a Computer-Aided Diagnosis framework the employed contrast limited adaptive histogram equalisation as a mechanism to enhance tumor regions in brain MRI images, 2-D Stationary Wavelet Transform to extract features, AdaBoost with a Support Vector Machine algorithms for normal and abnormal brain MRI classification achieving 99.45\% accuracy with 255 MRI images. In this paper, we have found in the existing literature that feature engineering, dimensionality reduction, and other hybrid approaches have been predominantly used to solve the brain MRI image classification problem. Various dimensionality reduction methods like Principal Component Analysis are employed to reduce the feature space of training datasets. Most recently, Talo et al. \cite{Talo2019} used Transfer Learning to perform binary classification of the brain, but our study proposes an expressive, fine-grain classifier that was trained on three distinct datasets. We have obtained an accuracy of 84.40\% on the National Institute of Neuroscience and Hospitals dataset which comprises 5,285 T1-weighted, 93.80\% on the Harvard Whole Brain Atlas comprising 1,133 T2-weighted images and 97.05\%  on the Biomedical School of Engineering which comprise 3,064 T1-weighted brain images using 5-fold cross-validation with ResNet50 CNN model.
\begin{table}[]
	\tiny
	\begin{tabular}{|p{1.30cm}|p{0.60cm}|p{0.60cm}|p{0.60cm}|p{0.60cm}|p{0.4cm}|p{1.7cm}|}
		\hline
		\textbf{Study} &\textbf{Accuracy Study I} &\textbf{Accuracy Study II} & \textbf{Accuracy Study III} & \textbf{Accuracy Study IV} & \textbf{Task} & \textbf{Method}\\
		\hline
		Cheng et al \cite{cheng2015correction}  & - & - & - & 91.28\% & Mutli & SVM and KNN \\
		\hline
		Paul et al\cite{paul2017deep}  &- & - & - & 91.43\% & Multi & CNN \\
		\hline
		Afshar et al \cite{afshar2019capsule} & - & - & - & 90.89\%  & Multi & CNN \\
		Anaraki et al\cite{anaraki2019magnetic} & - & - & 90.9\% & 94.2\% & Multi & GA-CNN \\
		Zacharaki et al\cite{zacharaki2009classification} & - & - & 85\% & - & Multi & SVM and KNN \\
		Zacharaki et al\cite{zacharaki2009classification} & - & - & 88\% & - & Binary & SVM and KNN \\
		El-Dahshan et al\cite{el2014computer} & - & - & 98\% & - & Binary & ANN and KNN \\
		Ertosum et al\cite{ertosun2015automated} & - & - & 71\% & - & Multi & CNN \\
		Ertosum et al\cite{ertosun2015automated} & - & - & 96\% & - & Binary & CNN \\
		Sultan et al\cite{sultan2019multi} & - & - & 98.7\% & 96.13\% & Multi & CNN \\
		Gudigar et al.\cite{gudigar2019application} & - & 97.38\% & - & - & Binary & PSO and SVM \\
		Zhang et al.\cite{zhang2015preclinical} & - & 99.33\% & - & - & Binary & GEPSVM \\
		Talo et al.\cite{Talo2019} & - & 100\% & - & - & Binary & Deep Transfer Learning \\
		\textbf{Proposed Study} & \textbf{85.23\%}& \textbf{93.80\%} & - & \textbf{97.05\%} & Multi & Deep Transfer Learning\\
		\hline
	\end{tabular}

	\caption{A comparative survey of brain MRI image classification methodologies}
	\label{tab:research_comparison}
\end{table}
In Fig. \ref{fig:system_deployment}, we have presented a proposed system deployment diagram depicting the infrastructural setup and configuration of our proposed framework.
\subsection{Conclusion and Future Direction}\label{conclusion}
identified the novelty of this research work that uses Deep Transfer Learning for multi-class classification of brain MRI images. Deep Transfer Learning demonstrates the potential of quickly adapting a model to solving a problem rather than building one from scratch and has shown to be suitable in terms of evaluation performance. Transfer Learning, as an active area of Machine Learning research, highlights the potential of accelerated adaptation of models across varied problem domains where underlying knowledge is invariant and transferable. We conducted three case studies using our novel dataset and two other publicly available datasets with separate contrast-enhancement techniques. Our research utilized data augmentation to address the limitations of the small dataset set size for the multi-class classification problem. We have shown the potential impact of Deep Transfer Learning as a potential research direction in solving the brain MRI multi-class classification task.
\bibliographystyle{IEEEtran}
\bibliography{IEEEfull,./Journal_Submisstion_V1}
\end{document}